\def\arxivbuild{}
\documentclass[sigconf,nonacm]{acmart}
\usepackage{xspace}
\usepackage{algorithm,algorithmic,tabularx,multirow,colortbl}
\usepackage{tcolorbox}
\usepackage{enumitem}
\setlist[itemize]{leftmargin=*,before={\emergencystretch=1.5em}}  %
\definecolor{gdBlue}{HTML}{3C5488}           %
\definecolor{gdBlueTint}{HTML}{EAF1FB}       %
\definecolor{gdTeal}{HTML}{00A087}           %
\definecolor{gdTealDark}{HTML}{00806C}       %
\definecolor{gdCyan}{HTML}{4DBBD5}           %
\definecolor{gdCyanDark}{HTML}{2B8FAB}       %
\definecolor{gdVermillion}{HTML}{E64B35}     %
\definecolor{gdVermillionDark}{HTML}{C4402B} %
\definecolor{gdSalmon}{HTML}{F39B7F}         %
\definecolor{gdBrown}{HTML}{7E6148}          %
\definecolor{gdLavender}{HTML}{8491B4}       %
\definecolor{gdLavenderDark}{HTML}{5E6A8E}   %
\definecolor{gdMint}{HTML}{91D1C2}           %
\definecolor{gdGreen}{HTML}{00806C}          %
\definecolor{gdInk}{HTML}{1E2329}            %
\definecolor{gdGray}{HTML}{5C636E}           %
\definecolor{gdGrayLight}{HTML}{C2C8D0}      %
\definecolor{gdGrayTint}{HTML}{F3F4F6}       %
\definecolor{gdRule}{HTML}{9AA3B5}           %
\newcommand{\ours}{\textsc{GraphDancer}\xspace}
\newif\ifsubmission
\submissiontrue
\ifdefined\arxivbuild \submissionfalse \fi

\setcopyright{none}
\renewcommand\footnotetextcopyrightpermission[1]{}
\acmDOI{}
\acmISBN{}
\acmYear{2027}
\copyrightyear{2027}
\acmConference[WWW '27]{The ACM Web Conference 2027}
  {May 10--14, 2027}{Dublin, Ireland}

\title[GraphDancer]{GraphDancer: Training LLMs to Explore and Reason over Graphs via Curriculum Reinforcement Learning}
\author{Yuyang Bai}
\affiliation{\institution{Texas A\&M University}\city{College Station}\state{TX}\country{USA}}
\email{ybai@tamu.edu}
\author{Zhuofeng Li}
\affiliation{\institution{Texas A\&M University}\city{College Station}\state{TX}\country{USA}}
\author{Ping Nie}
\affiliation{\institution{University of Waterloo}\city{Waterloo}\state{ON}\country{Canada}}
\author{Jianwen Xie}
\affiliation{\institution{Lambda}\country{USA}}
\author{Yu Wang}
\affiliation{\institution{University of Georgia}\city{Athens}\state{GA}\country{USA}}
\author{Yu Zhang}
\affiliation{\institution{Texas A\&M University}\city{College Station}\state{TX}\country{USA}}
\email{yuzhang@tamu.edu}
\renewcommand{\shortauthors}{Bai et al.}
\AtBeginDocument{\hypersetup{pdfauthor={Yuyang Bai, Zhuofeng Li, Ping Nie, Jianwen Xie, Yu Wang, Yu Zhang}}}

\begin{document}

\begin{abstract}
Large language models (LLMs) increasingly rely on external knowledge to
improve factuality, yet many knowledge sources on the Web, such as
scholarly indexes and product catalogs, are organized as heterogeneous
graphs.
Reasoning over such graph-structured knowledge poses two key challenges:
(1) navigating structured, schema-defined relations requires precise
function calls beyond similarity-based retrieval, and
(2) answering complex questions often demands multi-hop evidence aggregation
through iterative information seeking.
We study how LLMs can learn these interaction capabilities and transfer them to
unseen domains.
We propose \ours, a reinforcement learning (RL) framework that teaches LLMs
to navigate graphs by interleaving reasoning and function execution.
To improve RL for moderate-sized LLMs, we introduce a graph-aware
curriculum that schedules training by the structural difficulty of
questions using an easy-to-hard biased sampler.
We evaluate \ours on the multi-domain GRBench benchmark, training only on \textsc{Academic} and \textsc{Biomedical} graphs. Testing covers held-out questions from these domains as well as unseen \textsc{E-commerce}, \textsc{Literature}, and \textsc{Legal} graphs. With either Qwen2.5-3B-Instruct or Qwen3-14B as the backbone, \ours outperforms graph-agent baselines using the same backbone across all evaluation domains. These results indicate that the learned graph exploration and reasoning capabilities transfer across domains.
On average across unseen domains, \ours with a 14B backbone also outperforms baselines powered by the closed-source models GPT-4.1-mini and GPT-5-nano.
Trajectory analyses further show that \ours retrieves answer-bearing evidence more often and achieves higher accuracy when such evidence is found, while substantially reducing invalid-protocol outcomes.
Code is available \href{https://github.com/leopoldwhite/GraphDancer}{\textcolor{blue}{here}}.
\end{abstract}

\ccsdesc[500]{Computing methodologies~Reinforcement learning}
\ccsdesc[300]{Computing methodologies~Natural language processing}
\ccsdesc[300]{Computing methodologies~Knowledge representation and reasoning}
\ccsdesc[500]{Information systems~Question answering}
\keywords{Graph reasoning, knowledge graphs, language model agents, curriculum learning, reinforcement learning}
\maketitle

\section{Introduction}
\label{sec:introduction}

The internal parameters of large language models (LLMs) often capture
only part of external knowledge that is both rapidly
evolving~\cite{vu2024freshllms} and extremely long-tailed~\cite{sun2024head}.
Consequently, establishing a framework that enables LLMs to interact with
external knowledge sources can improve their factuality and
reliability~\cite{ji2023survey}.
Retrieval-augmented generation (RAG)~\cite{lewis2020retrieval} has become a
mainstream solution toward this goal. However, when external knowledge is
interconnected in graph
structures~\cite{han2024retrieval,sun2024think,ma2025think}, as in many
knowledge sources on the Web, from scholarly indexes and product catalogs
to heterogeneous knowledge graphs, two major challenges arise.

First, graphs often contain heterogeneous node and edge types (e.g.,
``[\textsc{User}] \textit{purchased} [\textsc{Product}]'' versus
``[\textsc{User}] \textit{viewed} [\textsc{Product}]''), so retrieval based solely on semantic similarity may confuse these relations. Instead, precisely defined function calls can
access the intended relation directly (e.g., finding products connected to a
given user via ``\textit{purchased}'').
Second, reasoning over graphs often requires capturing complex, multi-hop
connections between nodes. Such connections can make it difficult for the
model to acquire all necessary information in a single round, motivating an
adaptive, multi-round process that interleaves information seeking and
reasoning steps (e.g., subsequent function calls should be based on the product
nodes obtained in the current round).

To elicit multi-round graph interaction capabilities in LLMs, existing
studies~\cite{jin2024graph,amayuelas2025grounding,gao2025graph}
have proposed prompting strategies. However, these approaches rely
on the backbone's existing ability to follow the graph protocol and compose
successive operations. Compared with actual model training, they are less
effective at helping models internalize such graph interaction capabilities.
On the other hand, training models via supervised fine-tuning to
access external knowledge~\cite{asai2024self,li2026openresearcher} requires
examples of the desired interaction behavior and may struggle to generalize
to new graphs and domains, such as those with novel node/edge types,
function calls, or tasks.
Both gaps raise a question for accessing and reasoning over structured Web knowledge: how can LLM agents learn to interpret relation semantics, compose schema-defined graph operations, and transfer these capabilities to unseen domains? Such transfer would support knowledge-grounded reasoning over heterogeneous sources, allowing agents to work with unfamiliar schemas while reducing the need for source-specific training.

\begin{figure*}[t]
  \centering
  \includegraphics[width=\textwidth]{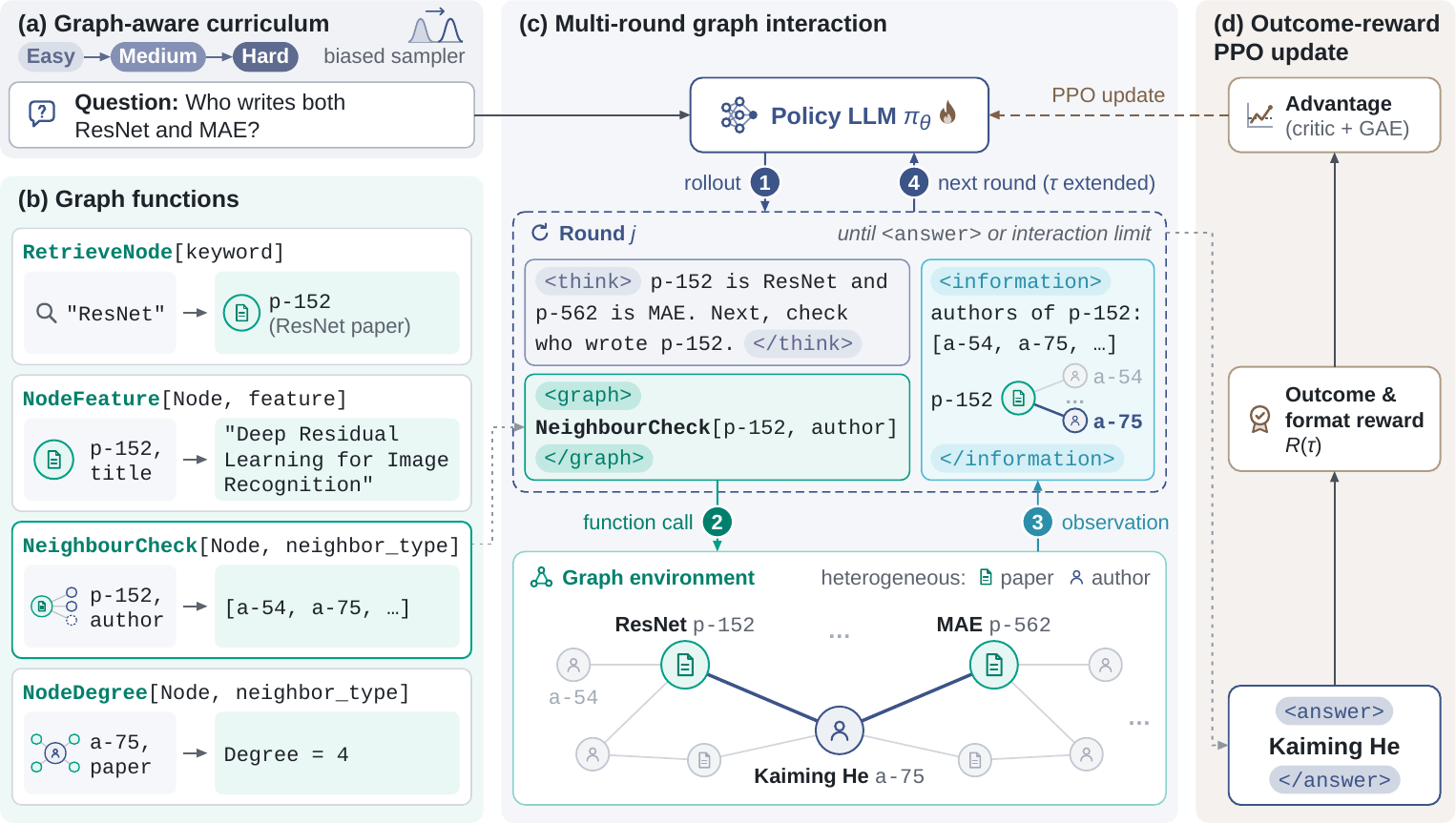}
  \caption{Overview of the \ours framework. (a)~The graph-aware curriculum
  shifts question sampling from easy to hard with a biased mixture. (b)~Graph
  functions with example inputs and outputs. (c)~In each round, the policy
  LLM reasons, calls one graph function, and receives the observation, until
  it answers or reaches the interaction limit. (d)~An outcome and format
  reward on the transcript drives the PPO update.}
  \Description{Four-panel overview. (a) The graph-aware curriculum
  (Easy, Medium, Hard, biased sampler) supplies the training question. (b) The four graph functions with an
  example input and output each. (c) One interaction round, in which the policy LLM
  reasons, issues a graph function call, and receives the observation from a
  heterogeneous graph environment, repeating until it answers or reaches the
  interaction limit. (d) The transcript receives an outcome and format
  reward, from which the advantage is estimated for the PPO update.}
  \label{fig:framework}
\end{figure*}

\smallskip
\noindent \textbf{Contributions.}
To address these challenges, we propose a reinforcement
learning (RL) framework, \ours{} (Figure~\ref{fig:framework}), that enables LLMs to learn to
interact with and reason over graphs in an adaptive, interleaved fashion.
The success of RL in enhancing LLM
reasoning~\cite{guo2025deepseek} and information
seeking~\cite{jin2025search} motivates this solution, as does the natural
formulation of graph interaction as a decision-making process with
graphs as \textbf{environments}, function calls as \textbf{actions}, and the
learning of graph interaction as \textbf{exploring the environment through actions}.
We optimize the policy with proximal policy optimization
(PPO)~\cite{schulman2017proximal}, using answer correctness and interaction
format as feedback. This trains the policy to use intermediate graph observations to decide
which operation to perform next and when to produce the final answer.

Meanwhile, to improve this solution for open-weight, moderate-sized
LLMs (e.g., Qwen2.5-3B-Instruct~\cite{hui2024qwen2} and Qwen3-14B~\cite{yang2025qwen3}), we incorporate curriculum
learning~\cite{bengio2009curriculum,parashar2025curriculum} into RL
training, progressively increasing task difficulty to mitigate their
inherent reasoning limitations compared to more advanced models.
Our \textbf{curriculum} introduces a \textbf{graph-aware} design
based on the structural difficulty of questions. At the beginning of RL
training, most sampled questions require only single-entity information
seeking over the graph (e.g., ``\textit{Who are the authors of the ResNet
paper?}''). As training progresses, exposure gradually shifts toward
questions involving multiple constraints, aggregation, or ranking over
multi-hop connections (e.g., ``\textit{Which venue did Kaiming He and Ross Girshick collaborate most?}''), while a biased-mixture sampler maintains
practice with simpler questions.

We train \ours on only two domains from the GRBench benchmark~\cite{jin2024graph},
\textsc{Academic} and \textsc{Biomedical}, and evaluate cross-domain
generalization on three \textbf{unseen domains}: \textsc{E-commerce},
\textsc{Literature}, and \textsc{Legal}. We also evaluate in-domain
performance on held-out questions from the two training domains.
The unseen-domain tests include \textbf{out-of-distribution (OOD)
question types} absent from training, such as questions requiring
inference beyond graph lookup while still drawing on graph context
(e.g., recommending a book based on a user's reading history).
Experimental results show that \ours consistently outperforms
graph-agent baselines~\cite{jin2024graph,gao2025graph} using the same
Qwen2.5-3B-Instruct or Qwen3-14B backbone. This advantage holds on
held-out in-domain questions and in every unseen domain. On average
across unseen domains, the 14B model also surpasses baselines powered
by the closed-source models GPT-4.1-mini~\cite{openai2025gpt41mini}
and GPT-5-nano~\cite{openai2025gpt5nano}. Controlled comparisons at
3B show that \ours outperforms vanilla PPO with uniform question
sampling across all four evaluation settings, supporting the value
of the graph-aware curriculum. Trajectory analyses reveal that
\ours retrieves answer-bearing evidence more often, achieves higher
answer accuracy when such evidence is found, and substantially
reduces invalid-protocol outcomes. These results indicate that
\ours learns transferable graph exploration and reasoning
capabilities while improving the reliability of multi-round graph
interaction.

The contributions of our work are as follows:
\begin{itemize}
\item Conceptually, we frame graph question answering as learning transferable capabilities for schema-defined knowledge access and multi-round reasoning.
\item Methodologically, we propose \ours, an RL framework with a graph-aware curriculum that combines easy-to-hard progression with continued practice across difficulty levels.
\item Empirically, we demonstrate cross-domain transfer and consistent gains over graph agents using the same backbone, with both 3B and 14B models.
\end{itemize}

\section{The \ours Framework}
\label{sec:method}

In this section, we present \ours, an RL framework that trains an LLM to
\emph{adaptively} interact with a graph via predefined function calls and
to \emph{interleave} information seeking with multi-step reasoning.
The graph environment executes each function call and returns its result,
while the policy decides which function to call next and when to answer
(Figure~\ref{fig:framework}). RL trains this multi-round interaction
behavior, guided by a \emph{graph-aware curriculum} that schedules training
questions by their structural difficulty. We define the task and the
interaction format in Sections~\ref{sec:setup} and~\ref{subsec:LLM_w_Graph},
the RL objective in Section~\ref{sec:ppo}, and the curriculum in
Sections~\ref{sec:difficulty} and~\ref{sec:curriculum}.

\subsection{Problem Setup}
\label{sec:setup}

Let $\mathcal G=(\mathcal V,\mathcal E)$ be a text-attributed graph with
heterogeneous node and edge types~\cite{jin2023heterformer}, such as an
academic network with papers, authors, and venues. Each node
$v\in\mathcal V$ carries a set of textual fields (e.g.,
\texttt{title}, \texttt{name}, \texttt{abstract}) and typed relations to
other nodes. Given a question $x$ with reference answer $y^\star$ and the
graph environment $\mathcal G$, the model must produce a natural-language
answer by interacting with $\mathcal G$ through a set of executable graph
functions.

We model question answering as an episodic Markov decision process whose
state holds the interaction history. At round $j$, the policy $\pi_\theta$
reads the question $x$, the graph schema $\mathcal S$, and its earlier
actions and observations. It generates an action $a_j$, which consists of
reasoning followed by one graph function call or the final answer. The
environment executes each call on $\mathcal G$ and returns an observation $o_j$,
\begin{equation}
 a_j\sim\pi_\theta(\cdot\mid x,\mathcal S,a_{<j},o_{<j}),\qquad
 o_j=\mathrm{Exec}_{\mathcal G}(a_j).
 \label{eq:loop}
\end{equation}
An episode terminates when the agent produces a final answer or reaches
the interaction limit.

\subsection{LLM Reasoning with Graph Interaction}
\label{subsec:LLM_w_Graph}

We use a simple interaction format to expose tool use to the LLM, inspired
by the multi-turn search calls of Search-R1~\cite{jin2025search}. The
interaction transcript (Figure~\ref{fig:framework}c) interleaves three block types:
\begin{itemize}
\item \emph{Reasoning block}: \texttt{<think>} $\cdots$
\texttt{</think>} containing free-text reasoning.
\item \emph{Action block}: \texttt{<graph>} $\cdots$
\texttt{</graph>} containing exactly one function call.
\item \emph{Observation block}: \texttt{<information>}
$\cdots$ \texttt{</information>} appended by the environment after
executing the action block.
\end{itemize}
The final answer goes in an \texttt{<answer>} $\cdots$
\texttt{</answer>} block.

\begin{algorithm}[tbp]
\small
\caption{Curriculum RL training of \ours}
\label{alg:training}
\begin{algorithmic}[1]
\REQUIRE Training questions $\mathcal D=\{(x,\mathcal G,y^\star)\}$, graph executor,
backbone, update budget $T=200$, batch size $B=128$
\STATE Initialize policy $\pi_\theta$ and critic $V_\phi$ from the backbone.
\STATE Assign structural levels (Section~\ref{sec:difficulty}) and form
shuffled queues $\mathcal D_0,\mathcal D_1,\mathcal D_2$.
\FOR{$t=0,\ldots,T-1$}
  \STATE Compute $p_t$ by Eqs.~\eqref{eq:gaussian}--\eqref{eq:curriculum}.
  \STATE Draw level counts $(n_0,n_1,n_2)\sim\mathrm{Multinomial}(B,p_t)$
  and take $n_k$ questions from each queue, reshuffling when exhausted.
  \FOR{each sampled $(x,\mathcal G,y^\star)$}
    \STATE Roll out agent tokens and execute graph calls to form
    $\tau$, stopping at an answer or the interaction limit.
    \STATE Record the agent-token mask, excluding observations
    and any supplied continuation prefix (Appendix~\ref{app:protocol}).
    \STATE Compute $R(\tau)$ by Eq.~\eqref{eq:reward} and assign
    it to the last generated token.
  \ENDFOR
  \STATE Compute masked GAE (Eq.~\eqref{eq:gae}) and critic returns,
  then normalize advantages over agent tokens.
  \STATE Update the critic on masked return errors and the policy
  with the PPO objective in Eq.~\eqref{eq:ppo} and entropy bonus,
  using one epoch.
\ENDFOR
\RETURN Policy $\pi_\theta$
\end{algorithmic}
\end{algorithm}

\smallskip
\noindent\textbf{Graph function calls.}
Standard RAG~\cite{lewis2020retrieval} retrieves free text. For
precise access over heterogeneous graphs, our agent invokes a small set of
predefined graph functions. Following the environment design
in Graph-CoT~\cite{jin2024graph}, the agent can call these graph functions
(Figure~\ref{fig:framework}b):
\begin{itemize}
\item \texttt{RetrieveNode[keyword]}: resolves a textual query to a
node identifier.
\item \texttt{NodeFeature[Node, feature]}:
returns a requested textual field of the node.
\item \texttt{NeighbourCheck[Node, neighbor\_type]}:
returns neighbor node identifiers under a specified typed relation.
\item \texttt{NodeDegree[Node, neighbor\_type]}: returns
the count of neighbors under a specified typed relation.
\end{itemize}
The environment validates and executes each function call and
returns its structured result (e.g., node identifiers or textual fields) or
error feedback in an observation block. Appendix~\ref{app:protocol} gives
the executable syntax and interaction limits.

\subsection{Training LLMs to Interact with Graphs}
\label{sec:ppo}

We train the policy $\pi_\theta$ with a rule-based reward
(Figure~\ref{fig:framework}d). The policy
samples only \emph{agent tokens} (reasoning, function calls, and the final
answer), while the graph environment executes function calls and \emph{injects
observation tokens}.
Let $z$ denote the concatenation of all agent-generated tokens in an
episode, and let $\tau=\mathrm{Env}(x,\mathcal G,z)$ be the full transcript
formed by interleaving $z$ with environment-injected observations. At
update $t$, we maximize the expected reward of $\tau$,
\begin{equation}
 J_t(\theta)=\mathbb E_{k\sim p_t,\,x\sim\mathcal D_k,\,
 z\sim\pi_\theta(\cdot\mid x,\mathcal S)}\bigl[R(\tau)\bigr],
 \label{eq:objective}
\end{equation}
where the curriculum sampler $p_t$ (Section~\ref{sec:curriculum}) picks a
level $k$, $\mathcal D_k$ holds the training questions of that level
(Section~\ref{sec:difficulty}), and $z$ is generated round by round as in
Eq.~\eqref{eq:loop}.
Observations stay in attention context, and only agent tokens enter the
policy and value losses.

\smallskip
\noindent\textbf{Reward design.}
We use a rule-based reward that prioritizes answer correctness and also rewards a valid interaction format. Let $C$ indicate whether the last complete answer exactly matches the reference after normalization, and let $F$ indicate whether the transcript's blocks are correctly ordered and closed. We also use $H$ to indicate whether an answer block can be extracted. Each indicator is 1 when its condition holds and 0 otherwise. The reward is
\begin{equation}
 R(\tau)=\alpha_C C+\alpha_F F+\alpha_H H(1-C)(1-F),
 \label{eq:reward}
\end{equation}
where $\alpha_C+\alpha_F=1$ and $\alpha_C>\alpha_F>\alpha_H>0$. These weights ensure that every correct answer receives a higher reward than every incorrect answer, with additional credit for valid formatting. When both the correctness and format checks fail, the $H$ term still provides a small reward if an answer block can be extracted. Appendix~\ref{app:optimization} lists the reward weights and normalization details.

\smallskip
\noindent\textbf{Policy optimization.}
We use PPO~\cite{schulman2017proximal} with a learned value function $V_\phi$ (the critic) and one rollout per question. We also experimented with GRPO~\cite{shao2024deepseekmath} but did not observe consistent performance gains over PPO in our setting. We therefore retain PPO throughout our curriculum comparisons, keeping the optimizer fixed to isolate the effect of question sampling. The critic estimates advantages across the multi-round interaction. At generated token positions $i$, where $s_i$ is the transcript prefix before agent token $z_i$, we propagate the terminal reward through generalized advantage estimation (GAE)~\cite{schulman2016high}:
\begin{equation}
 \delta_i=r_i+\gamma V_{\rm old}(s_{i+1})-V_{\rm old}(s_i),\qquad
 \widehat A_i=\delta_i+\gamma\lambda\widehat A_{i+1},
 \label{eq:gae}
\end{equation}
where $r_i$ equals $R(\tau)$ at the last generated token and zero elsewhere,
and $V_{\rm old}$ and $\pi_{\mathrm{old}}$ denote $V_\phi$ and $\pi_\theta$ before the update.
We use $\gamma=\lambda=1$ and a zero terminal value. The recurrence skips
observation positions. The critic fits returns
$V^{\rm targ}_i=\widehat A_i+V_{\rm old}(s_i)$ formed before advantage normalization.
For $\rho_i=\pi_\theta(z_i\mid s_i)/\pi_{\mathrm{old}}(z_i\mid s_i)$,
the policy surrogate is
\begin{equation}
 \ell_i=\min\!\left(\rho_i\widehat A_i,\,
 \operatorname{clip}(\rho_i,1-\epsilon,1+\epsilon)\widehat A_i\right).
 \label{eq:ppo}
\end{equation}
The objective adds a dual-clip~\cite{ye2020mastering} floor for negative
advantages and an entropy bonus, with reference-policy KL disabled.
Appendix~\ref{app:optimization} gives the clip range $\epsilon$, the critic
loss, and the other coefficients.

\subsection{Structural Difficulty of Graph Interaction}
\label{sec:difficulty}
\label{sec:round_types}

Questions over the same graph neighborhood can range from direct lookups to reasoning that combines constraints or compares candidates. We introduce a \emph{hierarchy of graph operations} that defines curriculum difficulty by how an answer must be constructed. We distinguish three classes of increasing order $c$. \emph{Single-anchor access} ($c=0$, \textbf{Easy}) retrieves an anchor's fields, neighbors, or degree (``\textit{Who are the authors of the ResNet paper?}''). \emph{Composition} ($c=1$, \textbf{Medium}) combines multiple anchors or constraints (``\textit{Who wrote both ResNet and MAE?}'' in Figure~\ref{fig:framework}). \emph{Set-level selection} ($c=2$, \textbf{Hard}) aggregates or ranks an expanded candidate set (``\textit{At which venue did Kaiming He and Ross Girshick collaborate most often?}''). These classes roughly correspond to lookup, join, and aggregation in database queries, capturing reasoning demands that hop count alone cannot express.

This hierarchy yields intuitive rules for labeling question templates. Words such as ``\textit{both}'' or additional conditions suggest composition, while ``\textit{most}'', ``\textit{least}'', or ``\textit{highest}'' suggest set-level selection. That said, the required operation matters more than the wording: ``\textit{How many papers did this author write?}'' remains Easy when \texttt{NodeDegree} directly supplies the count. For a training question $x$, we identify the graph operations $O(x)$ needed to construct an answer. We then assign the question to the level of its most demanding operation:
\begin{equation}
 d(x)=\max_{o\in O(x)}c(o)\in\{0,1,2\}.
 \label{eq:difficulty}
\end{equation}
For example, a question may require the agent to combine several constraints and then rank the resulting candidates. We label this question Hard because ranking requires set-level reasoning. We use these levels to group the training questions $\mathcal D$ into three pools, $\mathcal D_k=\{(x,\mathcal G,y^\star)\in\mathcal D:d(x)=k\}$. The curriculum described next controls how often questions are sampled from each pool during training. Appendix~\ref{app:labels} provides the detailed labeling rules.

\begin{table}[tbp]
\centering
\small
\setlength{\tabcolsep}{3pt}
\caption{GRBench questions by difficulty group and partition. In-domain
columns pool \textsc{Academic} and \textsc{Biomedical}.}
\label{tab:data_stats}
\begin{tabular*}{\linewidth}{@{\extracolsep{\fill}}lrrrrrrr@{}}
\toprule
 & \multicolumn{3}{c}{In-domain} & \multicolumn{3}{c}{Unseen-domain test} & \\
\cmidrule(lr){2-4}\cmidrule(lr){5-7}
Difficulty & Train & Val. & Test & \textsc{E-com.} & \textsc{Lit.} & \textsc{Legal} & All test \\
\midrule
Easy & 234 & 40 & 66 & 80 & 130 & 80 & 356 \\
Medium & 273 & 33 & 77 & 10 & 30 & 50 & 167 \\
Hard & 242 & 34 & 71 & 70 & 70 & 30 & 241 \\
OOD & 0 & 0 & 0 & 40 & 10 & 20 & 70 \\
\arrayrulecolor{gdRule}\midrule\arrayrulecolor{black}
Total & 749 & 107 & 214 & 200 & 240 & 180 & 834 \\
\bottomrule
\end{tabular*}
\vspace{-0.5em}
\end{table}

\begin{table*}[tp]
\centering
\small
\setlength{\tabcolsep}{4.4pt}
\caption{Main GRBench results, grouped by method. Results for TextRAG, GraphRAG, and Graph-CoT (GPT-3.5-turbo) are taken from~\cite{jin2024graph}, and those for Graph-MCTS and Graph-O1 from~\cite{liu2025grapho1}.
R-L: ROUGE-L (\%). GS: GPTScore (\%).
AVG: equal-weight mean across the four evaluation settings. Bold: best score per column.
$\Delta$\% reports the relative gain of \ours over each baseline, averaged across the evaluation settings reported for that baseline. Each gain is computed as $100(s^\ast-s)/s$, where $s^\ast$ is the \ours{} score and $s$ is the baseline score. R-L and GS gains are averaged separately.
--: unreported or inapplicable.}
\label{tab:main}
\begin{tabular}{llrrrrrrrrrrrr}
\toprule
 & & \multicolumn{2}{c}{In-domain} & \multicolumn{2}{c}{\textsc{E-commerce}} & \multicolumn{2}{c}{\textsc{Literature}} & \multicolumn{2}{c}{\textsc{Legal}} & \multicolumn{2}{c}{AVG} & \multicolumn{2}{c}{$\Delta$\%} \\
\cmidrule(lr){3-4}\cmidrule(lr){5-6}\cmidrule(lr){7-8}\cmidrule(lr){9-10}\cmidrule(lr){11-12}\cmidrule(lr){13-14}
Method & Backbone & R-L & GS & R-L & GS & R-L & GS & R-L & GS & R-L & GS & \multicolumn{1}{c}{R-L} & \multicolumn{1}{c}{GS} \\
\midrule
TextRAG~\cite{jin2024graph} & GPT-3.5-turbo & \textcolor{gdGray}{--} & \textcolor{gdGray}{--} & 14.06 & 20.00 & 10.04 & 20.83 & 18.14 & 23.89 & \textcolor{gdGray}{--} & \textcolor{gdGray}{--} & \textcolor{gdGreen}{$+304.9$} & \textcolor{gdGreen}{$+166.3$} \\
GraphRAG~\cite{jin2024graph} & GPT-3.5-turbo & \textcolor{gdGray}{--} & \textcolor{gdGray}{--} & 17.52 & 28.00 & 14.94 & 24.17 & 18.66 & 22.22 & \textcolor{gdGray}{--} & \textcolor{gdGray}{--} & \textcolor{gdGreen}{$+219.9$} & \textcolor{gdGreen}{$+134.9$} \\
Graph-MCTS~\cite{liu2025graphmcts} & Mixtral-8x7B-Instruct & \textcolor{gdGray}{--} & \textcolor{gdGray}{--} & 33.71 & 37.22 & 37.52 & 45.83 & 26.23 & 27.78 & \textcolor{gdGray}{--} & \textcolor{gdGray}{--} & \textcolor{gdGreen}{$+71.7$} & \textcolor{gdGreen}{$+61.6$} \\
Graph-O1~\cite{liu2025grapho1} & Mixtral-8x7B-Instruct & \textcolor{gdGray}{--} & \textcolor{gdGray}{--} & 34.61 & 38.76 & 38.42 & 45.83 & 27.36 & 32.20 & \textcolor{gdGray}{--} & \textcolor{gdGray}{--} & \textcolor{gdGreen}{$+66.2$} & \textcolor{gdGreen}{$+50.0$} \\
\arrayrulecolor{gdRule}\specialrule{\lightrulewidth}{\dimexpr\aboverulesep+0.70pt\relax}{\dimexpr\belowrulesep+0.70pt\relax}\arrayrulecolor{black}
\multirow{4}{*}{Graph-CoT~\cite{jin2024graph}} & GPT-3.5-turbo & \textcolor{gdGray}{--} & \textcolor{gdGray}{--} & 42.40 & 44.50 & 41.59 & 46.25 & 30.52 & 28.33 & \textcolor{gdGray}{--} & \textcolor{gdGray}{--} & \textcolor{gdGreen}{$+46.3$} & \textcolor{gdGreen}{$+52.4$} \\
 & GPT-4.1-mini & 33.09 & 39.72 & 32.82 & 46.50 & 34.91 & 50.42 & 49.85 & 56.67 & 37.67 & 48.33 & \textcolor{gdGreen}{$+52.7$} & \textcolor{gdGreen}{$+23.5$} \\
 & GPT-5-nano & 59.89 & 59.81 & 51.02 & \textbf{51.00} & \textbf{54.20} & 62.08 & 54.31 & 55.00 & 54.86 & 56.97 & \textcolor{gdGreen}{$+2.0$} & \textcolor{gdGreen}{$+2.9$} \\
 & Qwen3-14B & 50.94 & 50.47 & 47.68 & 48.00 & 49.63 & 61.25 & 45.90 & 51.11 & 48.54 & 52.71 & \textcolor{gdGreen}{$+15.4$} & \textcolor{gdGreen}{$+11.6$} \\
\arrayrulecolor{gdRule}\specialrule{\lightrulewidth}{\dimexpr\aboverulesep+0.70pt\relax}{\dimexpr\belowrulesep+0.70pt\relax}\arrayrulecolor{black}
\multirow{4}{*}{Graph Counselor~\cite{gao2025graph}} & GPT-3.5-turbo & 31.28 & 36.92 & 41.18 & 40.00 & 44.27 & 50.42 & 16.09 & 23.33 & 33.21 & 37.67 & \textcolor{gdGreen}{$+100.1$} & \textcolor{gdGreen}{$+68.3$} \\
 & GPT-4.1-mini & 36.82 & 46.26 & 24.64 & 30.00 & 29.97 & 45.42 & 30.82 & 33.89 & 30.56 & 38.89 & \textcolor{gdGreen}{$+85.2$} & \textcolor{gdGreen}{$+53.8$} \\
 & GPT-5-nano & 32.75 & 39.72 & 31.59 & 34.50 & 33.34 & 42.92 & 19.63 & 21.67 & 29.33 & 34.70 & \textcolor{gdGreen}{$+101.1$} & \textcolor{gdGreen}{$+80.6$} \\
 & Qwen3-14B & 53.41 & 59.35 & 38.29 & 41.00 & 49.75 & 54.58 & 21.44 & 26.67 & 40.72 & 45.40 & \textcolor{gdGreen}{$+56.5$} & \textcolor{gdGreen}{$+41.2$} \\
\arrayrulecolor{gdRule}\specialrule{\lightrulewidth}{\dimexpr\aboverulesep+0.70pt\relax}{\dimexpr\belowrulesep+0.70pt\relax}\arrayrulecolor{black}
\rowcolor{gdBlueTint}
\ours & Qwen3-14B & \textbf{61.32} & \textbf{62.62} & \textbf{51.62} & 50.00 & 53.09 & \textbf{62.50} & \textbf{57.80} & \textbf{59.44} & \textbf{55.96} & \textbf{58.64} & \textcolor{gdGray}{--} & \textcolor{gdGray}{--} \\
\bottomrule
\end{tabular}
\end{table*}

\subsection{Graph-Aware Curriculum RL}
\label{sec:curriculum}

RL on complex questions can provide sparse correctness feedback when moderate-sized LLMs perform poorly at the start of training~\cite{parashar2025curriculum}. The hierarchy in Section~\ref{sec:difficulty} suggests beginning with basic graph access and gradually emphasizing composition and set-level reasoning.
Hard questions still depend on basic graph access: before aggregating or
ranking candidates, the agent must identify the relevant entities and
follow the intended relations. An error at this stage can leave later
reasoning without the evidence it needs. Advancing to harder questions
therefore does not remove the need to practice these shared operations.
We propose a \textit{biased-mixture sampler} that increases exposure to demanding
questions while explicitly maintaining practice across difficulty levels.

Let $k$ index the $K=3$ difficulty levels and $u_t=t/(T-1)$ denote progress through $T$ RL updates. Following Gaussian curriculum sampling~\cite{parashar2025curriculum}, we define a moving distribution with center $\mu_t=\mu(u_t)$ and width $\sigma>0$:
\begin{equation}
 g_t(k)=
 \frac{\exp[-(k-\mu_t)^2/(2\sigma^2)]}
 {\sum_{k'=0}^{K-1}\exp[-(k'-\mu_t)^2/(2\sigma^2)]}.
\label{eq:gaussian}
\end{equation}
The non-decreasing pacing function $\mu$ moves the center from $0$ to $K-1$, shifting the sampling focus \textbf{from Easy toward Hard}. We extend this moving Gaussian with a fixed coverage distribution $q$ that assigns positive probability to every level. The resulting sampling distribution is
\begin{equation}
 p_t(k)=(1-\eta_t)g_t(k)+\eta_t q(k),
 \qquad \eta_t=\eta(u_t)\in[0,1].
\label{eq:curriculum}
\end{equation}
The Gaussian component controls the current focus of training. The prior $q$ specifies how continued practice is distributed across difficulty levels, while $\eta_t$ determines how much weight this practice receives in the overall sampler. This makes \textbf{continued practice} an explicit part of the curriculum, beyond the tail probabilities of the moving Gaussian. Since $p_t(k)\geq\eta_t q(k)$, every level has a lower bound on its expected share of training questions, which is positive whenever $\eta_t>0$. We increase $\eta_t$ during training to raise this coverage floor as the Gaussian moves toward harder levels. Setting $\eta_t=0$ recovers pure Gaussian sampling, while $\eta_t=1$ samples only from $q$. At each update, we draw level counts from $\operatorname{Multinomial}(B,p_t)$ for a batch of $B$ questions, then take questions from shuffled per-level queues, reshuffling each queue when exhausted. Appendix~\ref{app:optimization} gives the pacing, mixing, and coverage settings, and Algorithm~1 summarizes the training procedure.

\section{Experiments}
\label{sec:exp}

\subsection{Experimental Setup}
\label{sec:exp_setup}

\noindent\textbf{Dataset.}
We use GRBench~\cite{jin2024graph}, which provides executable graph environments across domains with a shared tool interface. We train on \textsc{Academic} and \textsc{Biomedical}, evaluating held-out in-domain questions and transfer to unseen \textsc{E-commerce}, \textsc{Literature}, and \textsc{Legal} domains. The 1,070-question in-domain pool follows a fixed 70:10:20 training/validation/test split, with graph-level quotas and question-template stratification. These partitions share the underlying graphs. The three unseen domains contribute 620 additional questions, giving 834 test questions in total. In-domain scores pool both domains with equal weight per question. Table~\ref{tab:data_stats} summarizes the partition.

Our \emph{Easy}, \emph{Medium}, and \emph{Hard} labels follow the operation hierarchy in Section~\ref{sec:difficulty}, independently of GRBench's original difficulty categories. Beyond domain transfer, we explicitly evaluate generalization to \emph{unseen question types}. The unseen-domain tests include 70 OOD questions whose types are excluded from training. These correspond to GRBench's original \emph{hard} category and include tasks requiring inference beyond graph lookup, such as book recommendation. We analyze OOD separately from the three structural difficulty levels. Appendix~\ref{app:labels} details the labeling procedure.

\smallskip
\noindent\textbf{Baselines.}
Table~\ref{tab:main} groups the systems by method. With Qwen2.5-3B-Instruct and Qwen3-14B, we compare \ours against Graph-CoT~\cite{jin2024graph}, which interleaves reasoning and graph function calls, and Graph Counselor~\cite{gao2025graph}, which coordinates multiple agents. We also report both methods with GPT-3.5-turbo~\cite{openai2023gpt35turbo}, GPT-4.1-mini~\cite{openai2025gpt41mini}, and GPT-5-nano~\cite{openai2025gpt5nano}. Graph-MCTS~\cite{liu2025graphmcts} and Graph-O1~\cite{liu2025grapho1} both use Mixtral-8x7B-Instruct~\cite{jiang2024mixtral} for search and reward-guided graph exploration. TextRAG and GraphRAG~\cite{jin2024graph} both use GPT-3.5-turbo to generate a single answer from retrieved node text. GraphRAG adds linearized one-hop neighborhoods.

\smallskip
\noindent\textbf{Implementation.}
We train Qwen2.5-3B-Instruct~\cite{hui2024qwen2} and
Qwen3-14B~\cite{yang2025qwen3} for 200 PPO updates of 128
trajectories each, with actor and critic learning rates of $10^{-6}$ and
$10^{-5}$, respectively. Both backbones use the same partition, difficulty
labels, reward, and curriculum (Section~\ref{sec:method}). Checkpoints are
selected on the validation partition.

Rollouts are sampled at temperature 0.7, top-$p$ 0.8, and top-$k$ 20.
Generations are capped at 500 tokens and observations at 1,024, with up to
ten rounds and a final generation with tool execution disabled.
Rolling contexts hold up to 24,576 tokens in training and 30,000
in evaluation. Qwen3 keeps its thinking mode after observations.
Appendix~\ref{app:implementation} gives the full prompt and settings.

\smallskip
\noindent\textbf{Evaluation metrics.}
We report two complementary metrics.
ROUGE-L (R-L)~\cite{lin2004rouge,lin2004automatic} compares the generated answer to
the reference, and GPTScore (GS)~\cite{jin2024graph} gives the percentage of answers that an LLM
judge marks correct under the GRBench evaluation prompt. AVG equally weights the four evaluation settings.

\begin{figure}[tbp]
\centering
\includegraphics[width=\linewidth]{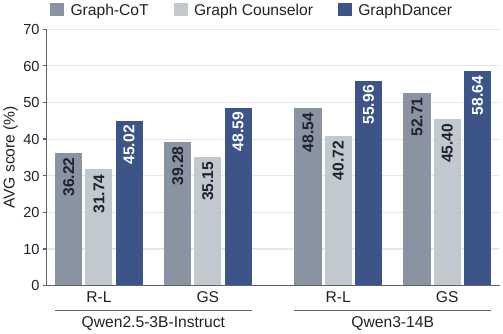}
\caption{AVG R-L and GS with each Qwen backbone. Graph-CoT and
Graph Counselor use the models before our RL training.}
\Description{Grouped bar chart of AVG R-L and GS (\%) for Graph-CoT,
Graph Counselor, and GraphDancer. With Qwen2.5-3B-Instruct, R-L is 36.22,
31.74, and 45.02, and GS is 39.28, 35.15, and 48.59. With Qwen3-14B, R-L is
48.54, 40.72, and 55.96, and GS is 52.71, 45.40, and 58.64.}
\label{fig:avg_scores}
\vspace{-0.5em}
\end{figure}

\subsection{Overall Performance}
\label{sec:main_results}

In Figure~\ref{fig:avg_scores}, the 3B and 14B \ours policies achieve 45.02 and
55.96 average R-L, respectively. Per domain column, the 14B policy gains
2.0\% R-L on average over the strongest reference row, Graph-CoT with GPT-5-nano,
and 15.4\% and 56.5\% over the same-backbone Graph-CoT and Graph Counselor
($\Delta$\% in Table~\ref{tab:main}). With either
Qwen backbone, \ours exceeds both same-backbone Graph-CoT and Graph
Counselor in R-L on all four settings. The gains on unseen domains suggest that the trained policies can apply
graph exploration and reasoning capabilities to new graph content and
schemas under the shared tool interface. Detailed 3B results are provided in
Appendix~\ref{app:analysis}.

At 3B, \ours exceeds every reported R-L AVG of the closed-model
graph-agent rows with GPT-3.5-turbo or GPT-4.1-mini (45.02 versus at most
37.67) and exceeds Graph-CoT with GPT-3.5-turbo in each reported column. At
14B, \ours exceeds all of these rows in every column. Against the
GPT-5-nano rows, \ours at 14B has a higher R-L AVG than Graph-CoT (55.96
versus 54.86), and \ours exceeds Graph Counselor at both sizes in every
column (R-L AVG 45.02 and 55.96 versus 29.33). Among the reference rows, the single-pass TextRAG and GraphRAG
score below every multi-round graph agent on \textsc{E-commerce} and
\textsc{Literature}, supporting interactive graph access.

GPTScore supports the same comparisons within each backbone. \ours has the
highest GS among the same-backbone systems in every column at both sizes, averaging 48.59 at 3B and
58.64 at 14B, versus at most 39.28 and 52.71 for the two baselines. At 3B, its GS average exceeds that of the stronger
GPT-4.1-mini row, Graph-CoT with GPT-4.1-mini. At 14B, \ours has a higher
GS than every GPT-3.5-turbo and GPT-4.1-mini row in each column. Against
the GPT-5-nano rows, its GS average at 14B is higher than that of Graph-CoT
(58.64 versus 56.97), and at both sizes it is higher than that of Graph
Counselor (48.59 and 58.64 versus 34.70).

\subsection{Impact of the Graph-Aware Curriculum}
\label{sec:curriculum_analysis}

To isolate the graph-aware curriculum, we train Qwen2.5-3B-Instruct
with PPO under other question-sampling schedules
(Table~\ref{tab:curriculum}). All policies share the backbone, reward,
optimizer, 200-update budget, validation-based checkpoint selection, and
test protocol. \emph{Uniform sampling.} Vanilla PPO keeps the
hyperparameters of \ours and replaces the sampler $p_t$ in
Eq.~\eqref{eq:curriculum} with uniform sampling over the same 749 training
questions. The other four schedules each change one design choice of
$p_t$. \emph{Order.} Static
Mixture fixes $p_t$ at its mean over the $T$ updates, which keeps the
expected cumulative level exposure of \ours and removes the progression.
Hard-to-Easy samples from $p_{T-1-t}$, which keeps this exposure and
reverses the direction. \emph{Pace.} Fast Pacing traverses $p_t$ in 100
updates and then holds $p_{T-1}$, which doubles the pace and adds Hard
exposure. \emph{Mixture.} Pure E2H keeps only the easy-to-hard Gaussian
$g_t$ (Eq.~\eqref{eq:gaussian}) by setting $\eta_t=0$.

\begin{table}[!t]
\centering
\small
\setlength{\tabcolsep}{3.12pt}
\caption{R-L (\%) on the test partition for the question-sampling
schedules of Section~\ref{sec:curriculum_analysis}, each trained with PPO
on Qwen2.5-3B-Instruct. In-domain: pooled. AVG: equal-weight mean.
Bold: best per column.}
\label{tab:curriculum}
\begin{tabular}{lrrrrr}
\toprule
Method & In-domain & \textsc{E-commerce} & \textsc{Literature} & \textsc{Legal} & AVG \\
\midrule
Vanilla PPO & 42.81 & 44.86 & 45.65 & 34.97 & 42.07 \\
\arrayrulecolor{gdRule}\midrule\arrayrulecolor{black}
Static Mixture & 37.83 & 41.99 & 43.19 & 28.37 & 37.85 \\
Hard-to-Easy & 42.31 & 45.85 & 44.14 & 29.50 & 40.45 \\
Fast Pacing & 39.09 & 44.90 & 45.89 & 36.77 & 41.66 \\
Pure E2H & 39.48 & 46.59 & \textbf{48.25} & 35.04 & 42.34 \\
\arrayrulecolor{gdRule}\midrule\arrayrulecolor{black}
\rowcolor{gdBlueTint}
\ours & \textbf{45.65} & \textbf{48.92} & 46.89 & \textbf{38.62} & \textbf{45.02} \\
\bottomrule
\end{tabular}
\vspace{-0.5em}
\end{table}

\begin{table*}[tbp]
\centering
\small
\setlength{\tabcolsep}{3.91pt}
\caption{Difficulty-wise R-L (\%) on the test partition. OOD keeps
GRBench's original \emph{hard} question types, all in the unseen domains,
separate from our Easy/Medium/Hard levels. All cells
average unstemmed per-question ROUGE-L. Bold: best per column within
each backbone.
Counts: Table~\ref{tab:data_stats}.}
\label{tab:difficulty_results}
\begin{tabular}{lrrrrrrrrrrrrrrr}
\toprule
 & \multicolumn{3}{c}{In-domain} & \multicolumn{4}{c}{\textsc{E-commerce}} & \multicolumn{4}{c}{\textsc{Literature}} & \multicolumn{4}{c}{\textsc{Legal}} \\
\cmidrule(lr){2-4}\cmidrule(lr){5-8}\cmidrule(lr){9-12}\cmidrule(lr){13-16}
Method & Easy & Medium & Hard & Easy & Medium & Hard & OOD & Easy & Medium & Hard & OOD & Easy & Medium & Hard & OOD \\
\midrule
Graph-CoT (3B) & 59.26 & 23.88 & 7.49 & 82.50 & \textbf{3.90} & 27.10 & \textbf{4.94} & 60.51 & \textbf{4.81} & 22.38 & 2.22 & 65.58 & 6.32 & 1.42 & \textbf{6.43} \\
Graph Counselor (3B) & 57.24 & 22.83 & 5.74 & 77.59 & 1.17 & 28.73 & 4.66 & 61.73 & 2.15 & 27.70 & \textbf{3.45} & 32.92 & 0.00 & 0.00 & 3.92 \\
\rowcolor{gdBlueTint}
\ours{} (3B) & \textbf{72.57} & \textbf{39.05} & \textbf{27.77} & \textbf{87.16} & 2.74 & \textbf{37.23} & 4.42 & \textbf{64.11} & 1.62 & \textbf{40.82} & 1.33 & \textbf{71.13} & \textbf{19.88} & \textbf{6.17} & 4.11 \\
\arrayrulecolor{gdRule}\midrule\arrayrulecolor{black}
Graph-CoT (14B) & 71.00 & 54.00 & 28.97 & 83.75 & 0.00 & 37.94 & 4.49 & 66.90 & 6.38 & 43.07 & 0.87 & 71.83 & 37.98 & \textbf{3.85} & 25.04 \\
Graph Counselor (14B) & 71.53 & 57.95 & \textbf{31.65} & 65.13 & 0.00 & 32.08 & 4.88 & 66.58 & \textbf{17.60} & 38.77 & \textbf{2.03} & 47.50 & 0.00 & 0.67 & 2.00 \\
\rowcolor{gdBlueTint}
\ours{} (14B) & \textbf{92.42} & \textbf{62.86} & 30.73 & \textbf{88.00} & 0.00 & \textbf{43.67} & \textbf{5.70} & \textbf{70.65} & 8.55 & \textbf{47.14} & 0.00 & \textbf{86.44} & \textbf{55.86} & 1.33 & \textbf{32.84} \\
\bottomrule
\end{tabular}
\end{table*}

\smallskip
\noindent\textbf{The graph-aware curriculum yields higher \mbox{R-L} than uniform sampling in every column.}
In Table~\ref{tab:curriculum}, \ours has higher R-L than Vanilla PPO
in-domain (45.65 versus 42.81) and on \textsc{E-commerce}
(48.92 versus 44.86), \textsc{Literature} (46.89 versus 45.65), and \textsc{Legal}
(38.62 versus 34.97), averaging 45.02 versus 42.07, with gains from 1.24
points on \textsc{Literature} to 4.06 on \textsc{E-commerce}.

\smallskip
\noindent\textbf{Every schedule that changes one design choice has a lower average \mbox{R-L} than \ours.}
Static Mixture and Hard-to-Easy share the
expected level exposure of \ours, so their gaps to \ours reflect the order
of that exposure. Fast Pacing keeps the order and the final mixture of
\ours, so its gap reflects the faster pace and the added Hard exposure.
Removing the biased mixture lowers the average R-L from
45.02 (\ours) to 42.34 (Pure E2H), consistent with the mixture's role of
maintaining exposure to each level (Section~\ref{sec:curriculum}). Pure E2H
averages close to Vanilla PPO and scores lower in-domain.
These results support combining gradual easy-to-hard progression with
the biased mixture.

\subsection{Difficulty-Wise Analysis}
\label{sec:difficulty_analysis}
\label{sec:behavior}

Table~\ref{tab:difficulty_results} breaks down R-L by structural difficulty.
At both model sizes, \ours scores above both same-backbone baselines on
every Easy subset. At 3B, \ours scores highest on Hard
questions in all four evaluation settings, with the largest gain in-domain
(27.77 versus 7.49 for Graph-CoT). It also scores highest on Medium
questions in-domain and on \textsc{Legal}. At 14B, it leads on Hard
questions on \textsc{E-commerce} (43.67 versus 37.94) and
\textsc{Literature} (47.14 versus 43.07), and on Medium questions
in-domain (62.86 versus 57.95) and on \textsc{Legal} (55.86 versus 37.98).
On OOD question types at 14B, \ours scores highest on \textsc{Legal}
(32.84 versus 25.04) and \textsc{E-commerce}.

\subsection{Case Study}
\label{sec:case_study}

\providecommand{\gdcheck}{\begin{pgfpicture}\pgfsetlinewidth{1pt}\pgfsetroundcap\pgfsetroundjoin
  \pgfsetstrokecolor{gdTeal}\pgfpathmoveto{\pgfpoint{0pt}{3pt}}\pgfpathlineto{\pgfpoint{2.3pt}{0.6pt}}
  \pgfpathlineto{\pgfpoint{6.2pt}{6pt}}\pgfusepath{stroke}\end{pgfpicture}}
\providecommand{\gdcross}{\begin{pgfpicture}\pgfsetlinewidth{1pt}\pgfsetroundcap
  \pgfsetstrokecolor{gdVermillion}\pgfpathmoveto{\pgfpoint{0.5pt}{0.6pt}}\pgfpathlineto{\pgfpoint{5.3pt}{5.4pt}}
  \pgfpathmoveto{\pgfpoint{0.5pt}{5.4pt}}\pgfpathlineto{\pgfpoint{5.3pt}{0.6pt}}\pgfusepath{stroke}\end{pgfpicture}}
\providecommand{\gddots}{...}
\providecommand{\gdline}[1]{\par\hangindent=1.2em\hangafter=1 {\ttfamily\hyphenchar\font=-1 #1}\par}
\providecommand{\gdbad}[1]{\textcolor{gdVermillionDark}{#1}}
\providecommand{\gdkey}[1]{\textcolor{gdTealDark}{\bfseries #1}}
\providecommand{\gdsep}{\par\vspace{3.5pt}{\color{gdInk}\hrule height 0.4pt}\vspace{3.5pt}}
\begin{figure}[tbp]
\centering
\begin{tcolorbox}[
    colback=gdCyanDark!9!white,
    colframe=gdCyanDark!50!white,
    boxrule=0.8pt,
    arc=1.6mm,
    boxsep=0.4mm,
    left=0.9mm,
    right=0.5mm,
    top=0.8mm,
    bottom=0.8mm,
    after skip=0pt
    ]
\footnotesize\raggedright

\par\noindent \textbf{Question} (\textsc{E-commerce}): {\ttfamily\hyphenchar\font=-1 What is the usual price of the items often purchased together with Mugen Power 2750mAh Extended Battery for Samsung Galaxy S4 I9500 / I9505?}\par
\gdsep
\par\textbf{Vanilla PPO} (3B):\par
\gdline{\gdbad{<think>\gddots{} the graph structure provided does not contain information about items often purchased together. \gddots{}</think>}}
\gdline{\gdbad{<graph>NeighbourCheck[Mugen Power 2750mAh Extended Battery for Samsung Galaxy S4 I9500 / I9505, bought\_together\_item]</graph>}}
\gdline{\gdbad{<information>\gddots{}does not exist in the graph. \gddots{}</information>}}
\gdline{<think>Since the neighbor type \textquotesingle{}bought\_together\_item\textquotesingle{} does not exist in the graph, it\textquotesingle{}s not possible to determine \gddots{}</think>}
\gdline{<answer>Not possible to determine</answer> \hspace{0.4em}{\rmfamily\gdcross\,\textcolor{gdVermillionDark}{\textbf{Wrong}}}}
\gdsep
\par\textbf{\ours} (3B):\par
\gdline{<think>\gddots{} I need to first identify the item node for Mugen Power\gddots{}, then find its bought\_together\_item neighbors, and finally check the price feature of these neighbors.</think>}
\gdline{\gdkey{<graph>RetrieveNode[Mugen Power 2750mAh\gddots{}]</graph>}}
\gdline{<information>\gddots{}B00CXOD6BQ.</information>}
\gdline{<think>\gddots{} Next, I need to find the items that are bought together with this item, \gddots{}</think>}
\gdline{\gdkey{<graph>NeighbourCheck[B00CXOD6BQ, bought\_together\_item]</graph>}}
\gdline{<information>\gddots{}: [\textquotesingle{}B00BW0XZO0\textquotesingle{}].</information>}
\gdline{<graph>NodeFeature[B00BW0XZO0, price]</graph>}
\gdline{<information>\gddots{}: 44.99.</information>}
\gdline{<answer>44.99</answer> \hspace{0.4em}{\rmfamily\gdcheck\,\textcolor{gdTealDark}{\textbf{Correct}}}}
\end{tcolorbox}
\caption{Case study. Vanilla PPO calls \texttt{NeighbourCheck} on the product name, while \ours first resolves the name to a node ID with \texttt{RetrieveNode}. Red marks the round with the failed call and teal the decisive calls.\ifsubmission\else{}
Complete trajectories appear in Appendix~\ref{app:full_case_study}.\fi}
\label{fig:case_study}
\Description{A rounded box shows an unseen E-commerce question and two tagged
trajectories with the Qwen2.5-3B-Instruct backbone. Vanilla PPO calls
NeighbourCheck with the product name, receives an error that the node or
neighbor type does not exist, and answers that the price cannot be determined.
GraphDancer retrieves the item node, finds its bought-together neighbor, reads
that neighbor's price, and answers 44.99, the gold answer.}
\vspace{-0.5em}
\end{figure}

We inspect whether \ours chains graph calls correctly on an unseen schema,
using an \textsc{E-commerce} question that asks for the
usual price of items often purchased together with a named product. It
requires the path \emph{Anchor Item} $\rightarrow$ \emph{Neighbor Item}
$\rightarrow$ \emph{Price} and is labeled Hard because its template
requires aggregation. In Figure~\ref{fig:case_study}, with the same 3B
backbone, Vanilla PPO calls \texttt{NeighbourCheck} with the product name
as the node ID. The environment reports a missing node or neighbor type,
and the policy answers ``Not possible to determine.''
\ours first retrieves the anchor node, then its single
bought-together neighbor, and then that neighbor's price, returning the
gold answer 44.99 with three graph calls versus one for Vanilla PPO.
\ours resolves the node ID before querying typed neighbors and attributes.

\subsection{Error Analysis of Graph Interaction}
\label{sec:behavior_analysis}

\begin{figure}[tbp]
\centering
\includegraphics[width=\linewidth]{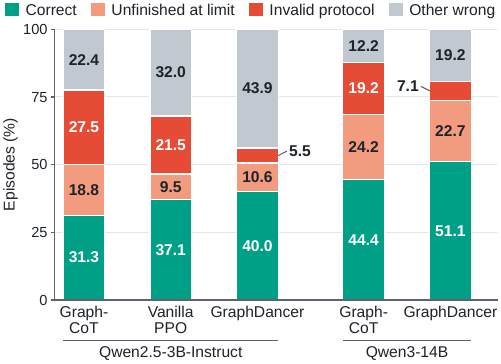}
\caption{Outcomes on all 834 test questions. Graph-CoT uses the model before
our RL training. Vanilla PPO and \ours apply uniform and curriculum
training, respectively. Each episode takes the first matching
category, checked from the bottom of each bar upward. Percentages label the
segments.}
\Description{Five vertical stacked bars in two groups. In the
Qwen2.5-3B-Instruct group, Graph-CoT has correct, unfinished-at-limit,
invalid-protocol, and other-wrong counts of 261, 157, 229, and 187, Vanilla
PPO has 309, 79, 179, and 267, and GraphDancer has 334, 88, 46, and 366. In
the Qwen3-14B group, Graph-CoT has 370, 202, 160, and 102, and GraphDancer
has 426, 189, 59, and 160, out of 834 questions each.}
\label{fig:error_breakdown}
\vspace{-0.5em}
\end{figure}

We distinguish answer correctness from invalid-protocol outcomes and
unfinished-at-limit episodes. Each episode is assigned to the first
matching category in the order correct (strict exact match, or strict
EM), unfinished at limit, invalid protocol (invalid format or call
syntax), and other wrong, so the categories are mutually exclusive.
Figure~\ref{fig:error_breakdown} separates two effects. Graph-CoT uses the
model before our RL training, so its comparison with the trained policies
shows the effect of training. Vanilla PPO and \ours share the 3B backbone,
reward, and interaction protocol, differing only in question sampling
(Section~\ref{sec:curriculum_analysis}).

At 3B, \ours answers 334 questions correctly, versus 261 for Graph-CoT,
and reduces invalid-protocol outcomes from 229 to 46 and
unfinished-at-limit episodes from 157 to 88. At 14B, it answers 426
questions correctly, versus 370 for Graph-CoT, and reduces invalid-protocol
outcomes from 160 to 59 and unfinished-at-limit episodes from 202 to 189.
Against Vanilla PPO at 3B, \ours raises correct answers from 309 to 334
and reduces invalid-protocol outcomes from 179 to 46. Curriculum sampling
thus reduces the invalid-protocol outcomes left by uniform training.
\ours also has the highest valid-format rate in every setting at both
sizes (Table~\ref{tab:vf}), based on the full tag-sequence check.

\begin{table}[tbp]
\centering
\small
\setlength{\tabcolsep}{5.6pt}
\caption{Valid-format rates (VF, \%) for single-agent protocols.
Bold: best per column within each backbone.}
\label{tab:vf}
\begin{tabular*}{\linewidth}{lrrrr}
\toprule
Method & In-domain & \textsc{E-commerce} & \textsc{Literature} & \textsc{Legal} \\
\midrule
\multicolumn{5}{@{}l}{\textcolor{gdGray}{\textit{Qwen2.5-3B-Instruct}}} \\
Graph-CoT & 32.24 & 60.00 & 46.25 & 47.22 \\
Vanilla PPO & 49.53 & 69.00 & 67.08 & 70.00 \\
\rowcolor{gdBlueTint}
\ours & \textbf{86.92} & \textbf{90.50} & \textbf{88.33} & \textbf{77.78} \\
\arrayrulecolor{gdRule}\midrule\arrayrulecolor{black}
\multicolumn{5}{@{}l}{\textcolor{gdGray}{\textit{Qwen3-14B}}} \\
Graph-CoT & 32.71 & 30.50 & 39.17 & 27.78 \\
\rowcolor{gdBlueTint}
\ours & \textbf{67.29} & \textbf{62.50} & \textbf{70.00} & \textbf{65.56} \\
\bottomrule
\end{tabular*}
\end{table}

\subsection{Evidence-Grounded Accuracy}
\label{sec:eh}

To test whether a policy retrieved the needed information, we complement
final-answer metrics with \emph{evidence hit} (EH), which checks whether
the normalized gold answer occurs in a returned tool observation. Each
observation is checked separately. $\mathrm{P}(\mathrm{EM}\mid\mathrm{EH})$
measures strict-EM correctness given a lexical hit.

\begin{table}[tbp]
\centering
\small
\setlength{\tabcolsep}{2.96pt}
\caption{Evidence-grounded accuracy (\%) pooled over 834 test questions.
EH: evidence hit. EM: strict exact match.
EM$\land$EH measures how often an answer is correct and has a hit.
The two conditional columns give EM after a hit and after a miss.
Bold: best per column within each backbone.}
\label{tab:interaction}
\label{tab:eh}
\begin{tabular}{lrrrrr}
\toprule
Method & EH & EM$\land$EH & $\mathrm P(\mathrm{EM}\mid\mathrm{EH})$
 & $\mathrm P(\mathrm{EM}\mid\neg\mathrm{EH})$ & EM \\
\midrule
\multicolumn{6}{l}{\textcolor{gdGray}{\textit{Qwen2.5-3B-Instruct}}} \\
Graph-CoT & 47.12 & 29.02 & 61.58 & 4.31 & 31.29 \\
Graph Counselor & 48.68 & 23.62 & 48.52 & 1.87 & 24.58 \\
\rowcolor{gdBlueTint}
\ours & \textbf{53.36} & \textbf{37.41} & \textbf{70.11} & \textbf{5.66} & \textbf{40.05} \\
\arrayrulecolor{gdRule}\midrule\arrayrulecolor{black}
\multicolumn{6}{l}{\textcolor{gdGray}{\textit{Qwen3-14B}}} \\
Graph-CoT & 60.31 & 42.21 & 69.98 & 5.44 & 44.36 \\
Graph Counselor & 58.39 & 35.37 & 60.57 & 7.78 & 38.61 \\
\rowcolor{gdBlueTint}
\ours & \textbf{63.91} & \textbf{47.72} & \textbf{74.67} & \textbf{9.30} & \textbf{51.08} \\
\bottomrule
\end{tabular}
\vspace{-0.5em}
\end{table}

At both sizes, \ours has higher EH, accuracy after a hit or a miss, and EM
than Graph-CoT and Graph Counselor on the same backbone (Table~\ref{tab:eh}). At 3B, it reaches 53.36\% EH versus 47.12\% and 48.68\%,
70.11\% accuracy after a hit versus 61.58\% and 48.52\%, and 40.05\% EM
versus 31.29\% and 24.58\%. At 14B, the corresponding values are 63.91\%
versus 60.31\% and 58.39\%, 74.67\% versus 69.98\% and 60.57\%, and
51.08\% versus 44.36\% and 38.61\%. \ours also achieves the highest
EM$\land$EH at both sizes (37.41\% and 47.72\%).
Together, the higher hit rate and conditional accuracy suggest
improvements in both retrieving answer-bearing information and answering
correctly after retrieval, extending the analysis beyond protocol compliance.
\section{Related Work}
\label{sec:related-work}

\noindent\textbf{Agentic RAG.}
Agentic RAG enables LLMs to interleave reasoning with information seeking
and adapt subsequent queries to newly gathered evidence.
Search-R1~\cite{jin2025search} and
R1-Searcher~\cite{song2025r1searcher} learn search policies through
outcome-based RL. WebDancer~\cite{wu2025webdancer} and
REDSearcher~\cite{chu2026redsearcher} develop staged training pipelines
for long-horizon search, while OpenResearcher~\cite{li2026openresearcher}
and OpenSeeker~\cite{du2026openseeker} emphasize synthesized
search-and-browse trajectories for supervised training.
These studies establish multi-round information seeking as a trainable
policy. \ours investigates this paradigm over heterogeneous graphs,
where evidence access requires composing schema-defined operations
over typed entities and relations.

\smallskip
\noindent\textbf{LLM agents on graphs.}
Think-on-Graph~\cite{sun2024think} and RoG~\cite{luo2024reasoning}
guide reasoning through entity-relation paths.
On GRBench, Graph-CoT~\cite{jin2024graph} interleaves reasoning with
graph function calls, while Graph Counselor~\cite{gao2025graph}
coordinates multiple agents. GraphRunner~\cite{kashmira2025graphrunner}
separates traversal planning, verification, and execution.
Graph-MCTS~\cite{liu2025graphmcts} uses Monte Carlo tree search to
evaluate candidate traversals, and Graph-O1~\cite{liu2025grapho1}
combines tree search with RL to improve exploration decisions.
Recent post-training methods also evaluate cross-domain transfer on
GRBench. GraphScout~\cite{ying2026graphscout} uses a strong model to
generate graph-grounded questions and evidence clues for RL training.
GraphCert~\cite{jiang2026graphcert} verifies supervision generated by a
compact local model and turns accepted evidence into certified rubrics
for policy optimization. These methods highlight the importance of
supervision quality and evidence-aware feedback.
Related approaches include Graph-R1~\cite{luo2025graph}, which applies
RL to multi-turn retrieval over knowledge hypergraphs, and
Structure-R1~\cite{wu2025structure}, which learns to organize retrieved
content into structured representations.
\ours focuses on how the order and continued coverage of training
questions shape transferable graph interaction capabilities.

\smallskip
\noindent\textbf{RL for reasoning and curriculum learning.}
RL has been used to align LLMs with human preferences through
PPO~\cite{ouyang2022training,schulman2017proximal}, to enhance
reasoning~\cite{guo2025deepseek}, and to train multi-turn
information seeking through outcome feedback~\cite{jin2025search}.
To ease learning in long-horizon tasks, prior work adds curriculum
learning~\cite{bengio2009curriculum} to RL through strategies such as
easy-to-hard scheduling~\cite{parashar2025curriculum,team2025kimi,bercovich2025llama},
adaptive rollout control~\cite{shen2025thinking}, and multi-objective
balancing~\cite{hammoud2025train}.
We adapt curriculum RL to the structure of graph interaction by
distinguishing single-anchor access, constraint composition, and
set-level selection. Our schedule gradually shifts its focus across these demands while
explicitly maintaining exposure to all levels, since harder questions
still depend on reliable basic graph access.

\section{Conclusions and Future Work}
\label{sec:conclusion}

\ours shows that graph-agent training benefits from organizing practice
around the operations needed to construct an answer, while maintaining
exposure across levels as reasoning demands increase.
Controlled comparisons under a fixed optimizer indicate that matching
cumulative difficulty exposure, reversing the progression, or removing
explicit coverage yields lower average performance.
These results suggest that training order matters alongside the
distribution of practice.
Evidence analyses distinguish improvements in reaching answer-bearing
observations from improvements in answering correctly after retrieval,
while error analyses show more reliable protocol execution.
Together, the findings suggest that continued practice with basic graph
access supports learning more complex compositions that transfer across
domains. Structural difficulty thus offers a practical basis for training
agents to retrieve and reason over heterogeneous Web knowledge.

Future work can refine several aspects of \ours. Our template-based
difficulty labels provide coarse estimates of question demands, while
the curriculum follows a predefined schedule. Finer difficulty estimates
and pacing adapted to per-level progress could improve training
efficiency. Further studies could examine sensitivity to curriculum
settings, schema descriptions, and demonstrations, while extending
evaluation to additional tool interfaces and noisier observations.
Rewards for evidence use could also provide more targeted feedback
during training.

\bibliographystyle{ACM-Reference-Format}
\bibliography{references}


\providecommand{\bibqmark}[1]{?\ifx.#1\else#1\fi}
\begin{thebibliography}{46}

%%% the Web address.

\ifx \showCODEN    \undefined \def \showCODEN     #1{\unskip}     \fi
\ifx \showISBNx    \undefined \def \showISBNx     #1{\unskip}     \fi
\ifx \showISBNxiii \undefined \def \showISBNxiii  #1{\unskip}     \fi
\ifx \showISSN     \undefined \def \showISSN      #1{\unskip}     \fi
\ifx \showLCCN     \undefined \def \showLCCN      #1{\unskip}     \fi
\ifx \shownote     \undefined \def \shownote      #1{#1}          \fi
\ifx \showarticletitle \undefined \def \showarticletitle #1{#1}   \fi
\ifx \showURL      \undefined \def \showURL       {\relax}        \fi
% invisible to TeX
\providecommand\bibfield[2]{#2}
\providecommand\bibinfo[2]{#2}
\providecommand\natexlab[1]{#1}
\providecommand\showeprint[2][]{arXiv:#2}

\bibitem[Amayuelas et~al\mbox{.}(2025)]%
        {amayuelas2025grounding}
\bibfield{author}{\bibinfo{person}{Alfonso Amayuelas}, \bibinfo{person}{Joy~Prakash Sain}, \bibinfo{person}{Simerjot Kaur}, {and} \bibinfo{person}{Charese Smiley}.} \bibinfo{year}{2025}\natexlab{}.
\newblock \showarticletitle{Grounding {LLM} Reasoning with Knowledge Graphs}. In \bibinfo{booktitle}{\emph{NeurIPS'25 Workshop on FoRLM}}.
\newblock


\bibitem[Asai et~al\mbox{.}(2024)]%
        {asai2024self}
\bibfield{author}{\bibinfo{person}{Akari Asai}, \bibinfo{person}{Zeqiu Wu}, \bibinfo{person}{Yizhong Wang}, \bibinfo{person}{Avirup Sil}, {and} \bibinfo{person}{Hannaneh Hajishirzi}.} \bibinfo{year}{2024}\natexlab{}.
\newblock \showarticletitle{{Self-RAG}: Learning to Retrieve, Generate, and Critique through Self-Reflection}. In \bibinfo{booktitle}{\emph{ICLR'24}}.
\newblock


\bibitem[Bengio et~al\mbox{.}(2009)]%
        {bengio2009curriculum}
\bibfield{author}{\bibinfo{person}{Yoshua Bengio}, \bibinfo{person}{J{\'e}r{\^o}me Louradour}, \bibinfo{person}{Ronan Collobert}, {and} \bibinfo{person}{Jason Weston}.} \bibinfo{year}{2009}\natexlab{}.
\newblock \showarticletitle{Curriculum Learning}. In \bibinfo{booktitle}{\emph{ICML'09}}. \bibinfo{pages}{41--48}.
\newblock


\bibitem[Chu et~al\mbox{.}(2026)]%
        {chu2026redsearcher}
\bibfield{author}{\bibinfo{person}{Zheng Chu}, \bibinfo{person}{Xiao Wang}, \bibinfo{person}{Jack Hong}, \bibinfo{person}{Huiming Fan}, \bibinfo{person}{Yuqi Huang}, \bibinfo{person}{Yue Yang}, \bibinfo{person}{Guohai Xu}, \bibinfo{person}{Chenxiao Zhao}, \bibinfo{person}{Cheng Xiang}, \bibinfo{person}{Shengchao Hu}, \bibinfo{person}{Dongdong Kuang}, \bibinfo{person}{Ming Liu}, \bibinfo{person}{Bing Qin}, {and} \bibinfo{person}{Xing Yu}.} \bibinfo{year}{2026}\natexlab{}.
\newblock \showarticletitle{{REDSearcher}: A Scalable and Cost-Efficient Framework for Long-Horizon Search Agents}.
\newblock \bibinfo{journal}{\emph{arXiv preprint arXiv:2602.14234}} (\bibinfo{year}{2026}).
\newblock


\bibitem[Du et~al\mbox{.}(2026)]%
        {du2026openseeker}
\bibfield{author}{\bibinfo{person}{Yuwen Du}, \bibinfo{person}{Rui Ye}, \bibinfo{person}{Shuo Tang}, \bibinfo{person}{Xinyu Zhu}, \bibinfo{person}{Yijun Lu}, \bibinfo{person}{Yuzhu Cai}, {and} \bibinfo{person}{Siheng Chen}.} \bibinfo{year}{2026}\natexlab{}.
\newblock \showarticletitle{{OpenSeeker}: Democratizing Frontier Search Agents by Fully Open-Sourcing Training Data}. In \bibinfo{booktitle}{\emph{COLM'26}}.
\newblock


\bibitem[Gao et~al\mbox{.}(2025)]%
        {gao2025graph}
\bibfield{author}{\bibinfo{person}{Junqi Gao}, \bibinfo{person}{Xiang Zou}, \bibinfo{person}{Ying Ai}, \bibinfo{person}{Dong Li}, \bibinfo{person}{Yichen Niu}, \bibinfo{person}{Biqing Qi}, {and} \bibinfo{person}{Jianxing Liu}.} \bibinfo{year}{2025}\natexlab{}.
\newblock \showarticletitle{{Graph Counselor}: Adaptive Graph Exploration via Multi-Agent Synergy to Enhance {LLM} Reasoning}. In \bibinfo{booktitle}{\emph{ACL'25}}. \bibinfo{pages}{24650--24668}.
\newblock


\bibitem[Guo et~al\mbox{.}(2025)]%
        {guo2025deepseek}
\bibfield{author}{\bibinfo{person}{Daya Guo}, \bibinfo{person}{Dejian Yang}, \bibinfo{person}{Haowei Zhang}, \bibinfo{person}{Junxiao Song}, \bibinfo{person}{Peiyi Wang}, \bibinfo{person}{Qihao Zhu}, \bibinfo{person}{Runxin Xu}, \bibinfo{person}{Ruoyu Zhang}, \bibinfo{person}{Shirong Ma}, \bibinfo{person}{Xiao Bi}, {et~al\mbox{.}}} \bibinfo{year}{2025}\natexlab{}.
\newblock \showarticletitle{{DeepSeek-R1} Incentivizes Reasoning in {LLMs} through Reinforcement Learning}.
\newblock \bibinfo{journal}{\emph{Nature}} \bibinfo{volume}{645}, \bibinfo{number}{8081} (\bibinfo{year}{2025}), \bibinfo{pages}{633--638}.
\newblock


\bibitem[Hammoud et~al\mbox{.}(2025)]%
        {hammoud2025train}
\bibfield{author}{\bibinfo{person}{Hasan Abed Al~Kader Hammoud}, \bibinfo{person}{Kumail Alhamoud}, \bibinfo{person}{Abed Hammoud}, \bibinfo{person}{Elie Bou-Zeid}, \bibinfo{person}{Marzyeh Ghassemi}, {and} \bibinfo{person}{Bernard Ghanem}.} \bibinfo{year}{2025}\natexlab{}.
\newblock \showarticletitle{Train Long, Think Short: Curriculum Learning for Efficient Reasoning}.
\newblock \bibinfo{journal}{\emph{arXiv preprint arXiv:2508.08940}} (\bibinfo{year}{2025}).
\newblock


\bibitem[Han et~al\mbox{.}(2024)]%
        {han2024retrieval}
\bibfield{author}{\bibinfo{person}{Haoyu Han}, \bibinfo{person}{Yu Wang}, \bibinfo{person}{Harry Shomer}, \bibinfo{person}{Kai Guo}, \bibinfo{person}{Jiayuan Ding}, \bibinfo{person}{Yongjia Lei}, \bibinfo{person}{Mahantesh Halappanavar}, \bibinfo{person}{Ryan~A. Rossi}, \bibinfo{person}{Subhabrata Mukherjee}, \bibinfo{person}{Xianfeng Tang}, {et~al\mbox{.}}} \bibinfo{year}{2024}\natexlab{}.
\newblock \showarticletitle{Retrieval-Augmented Generation with Graphs ({GraphRAG})}.
\newblock \bibinfo{journal}{\emph{arXiv preprint arXiv:2501.00309}} (\bibinfo{year}{2024}).
\newblock


\bibitem[Ji et~al\mbox{.}(2023)]%
        {ji2023survey}
\bibfield{author}{\bibinfo{person}{Ziwei Ji}, \bibinfo{person}{Nayeon Lee}, \bibinfo{person}{Rita Frieske}, \bibinfo{person}{Tiezheng Yu}, \bibinfo{person}{Dan Su}, \bibinfo{person}{Yan Xu}, \bibinfo{person}{Etsuko Ishii}, \bibinfo{person}{Ye~Jin Bang}, \bibinfo{person}{Andrea Madotto}, {and} \bibinfo{person}{Pascale Fung}.} \bibinfo{year}{2023}\natexlab{}.
\newblock \showarticletitle{Survey of Hallucination in Natural Language Generation}.
\newblock \bibinfo{journal}{\emph{Comput. Surveys}} \bibinfo{volume}{55}, \bibinfo{number}{12} (\bibinfo{year}{2023}), \bibinfo{pages}{1--38}.
\newblock


\bibitem[Jiang et~al\mbox{.}(2024)]%
        {jiang2024mixtral}
\bibfield{author}{\bibinfo{person}{Albert~Q. Jiang}, \bibinfo{person}{Alexandre Sablayrolles}, \bibinfo{person}{Antoine Roux}, \bibinfo{person}{Arthur Mensch}, \bibinfo{person}{Blanche Savary}, \bibinfo{person}{Chris Bamford}, \bibinfo{person}{Devendra~Singh Chaplot}, \bibinfo{person}{Diego de~las Casas}, \bibinfo{person}{Emma~Bou Hanna}, \bibinfo{person}{Florian Bressand}, {et~al\mbox{.}}} \bibinfo{year}{2024}\natexlab{}.
\newblock \showarticletitle{{Mixtral} of Experts}.
\newblock \bibinfo{journal}{\emph{arXiv preprint arXiv:2401.04088}} (\bibinfo{year}{2024}).
\newblock


\bibitem[Jiang et~al\mbox{.}(2026)]%
        {jiang2026graphcert}
\bibfield{author}{\bibinfo{person}{Weiqi Jiang}, \bibinfo{person}{Yuchen Ying}, \bibinfo{person}{Rui Wang}, \bibinfo{person}{Kaixuan Chen}, \bibinfo{person}{Bingde Hu}, \bibinfo{person}{Shunyu Liu}, \bibinfo{person}{Yu Wang}, {and} \bibinfo{person}{Tongya Zheng}.} \bibinfo{year}{2026}\natexlab{}.
\newblock \showarticletitle{{GraphCert}: Bootstrap Agentic Graph Reasoning with Certified Evidence Rubrics}.
\newblock \bibinfo{journal}{\emph{arXiv preprint arXiv:2609.38798}} (\bibinfo{year}{2026}).
\newblock


\bibitem[Jin et~al\mbox{.}(2024)]%
        {jin2024graph}
\bibfield{author}{\bibinfo{person}{Bowen Jin}, \bibinfo{person}{Chulin Xie}, \bibinfo{person}{Jiawei Zhang}, \bibinfo{person}{Kashob~Kumar Roy}, \bibinfo{person}{Yu Zhang}, \bibinfo{person}{Zheng Li}, \bibinfo{person}{Ruirui Li}, \bibinfo{person}{Xianfeng Tang}, \bibinfo{person}{Suhang Wang}, \bibinfo{person}{Yu Meng}, {and} \bibinfo{person}{Jiawei Han}.} \bibinfo{year}{2024}\natexlab{}.
\newblock \showarticletitle{{Graph Chain-of-Thought}: Augmenting Large Language Models by Reasoning on Graphs}. In \bibinfo{booktitle}{\emph{Findings of ACL'24}}. \bibinfo{pages}{163--184}.
\newblock


\bibitem[Jin et~al\mbox{.}(2025)]%
        {jin2025search}
\bibfield{author}{\bibinfo{person}{Bowen Jin}, \bibinfo{person}{Hansi Zeng}, \bibinfo{person}{Zhenrui Yue}, \bibinfo{person}{Jinsung Yoon}, \bibinfo{person}{Sercan Arik}, \bibinfo{person}{Dong Wang}, \bibinfo{person}{Hamed Zamani}, {and} \bibinfo{person}{Jiawei Han}.} \bibinfo{year}{2025}\natexlab{}.
\newblock \showarticletitle{{Search-R1}: Training {LLMs} to Reason and Leverage Search Engines with Reinforcement Learning}. In \bibinfo{booktitle}{\emph{COLM'25}}.
\newblock


\bibitem[Jin et~al\mbox{.}(2023)]%
        {jin2023heterformer}
\bibfield{author}{\bibinfo{person}{Bowen Jin}, \bibinfo{person}{Yu Zhang}, \bibinfo{person}{Qi Zhu}, {and} \bibinfo{person}{Jiawei Han}.} \bibinfo{year}{2023}\natexlab{}.
\newblock \showarticletitle{{Heterformer}: Transformer-Based Deep Node Representation Learning on Heterogeneous Text-Rich Networks}. In \bibinfo{booktitle}{\emph{KDD'23}}. \bibinfo{pages}{1020--1031}.
\newblock


\bibitem[Kashmira et~al\mbox{.}(2025)]%
        {kashmira2025graphrunner}
\bibfield{author}{\bibinfo{person}{Savini Kashmira}, \bibinfo{person}{Jayanaka~L. Dantanarayana}, \bibinfo{person}{Kriszti{\'a}n Flautner}, \bibinfo{person}{Lingjia Tang}, {and} \bibinfo{person}{Jason Mars}.} \bibinfo{year}{2025}\natexlab{}.
\newblock \showarticletitle{{GraphRunner}: A Multi-Stage Framework for Efficient and Accurate Graph-Based Retrieval}.
\newblock \bibinfo{journal}{\emph{arXiv preprint arXiv:2507.08945}} (\bibinfo{year}{2025}).
\newblock


\bibitem[{Kimi Team} et~al\mbox{.}(2025)]%
        {team2025kimi}
\bibfield{author}{\bibinfo{person}{{Kimi Team}}, \bibinfo{person}{Angang Du}, \bibinfo{person}{Bofei Gao}, \bibinfo{person}{Bowei Xing}, \bibinfo{person}{Changjiu Jiang}, \bibinfo{person}{Cheng Chen}, \bibinfo{person}{Cheng Li}, \bibinfo{person}{Chenjun Xiao}, \bibinfo{person}{Chenzhuang Du}, \bibinfo{person}{Chonghua Liao}, {et~al\mbox{.}}} \bibinfo{year}{2025}\natexlab{}.
\newblock \showarticletitle{{Kimi k1.5}: Scaling Reinforcement Learning with {LLMs}}.
\newblock \bibinfo{journal}{\emph{arXiv preprint arXiv:2501.12599}} (\bibinfo{year}{2025}).
\newblock


\bibitem[Lewis et~al\mbox{.}(2020)]%
        {lewis2020retrieval}
\bibfield{author}{\bibinfo{person}{Patrick Lewis}, \bibinfo{person}{Ethan Perez}, \bibinfo{person}{Aleksandra Piktus}, \bibinfo{person}{Fabio Petroni}, \bibinfo{person}{Vladimir Karpukhin}, \bibinfo{person}{Naman Goyal}, \bibinfo{person}{Heinrich K{\"u}ttler}, \bibinfo{person}{Mike Lewis}, \bibinfo{person}{Wen-tau Yih}, \bibinfo{person}{Tim Rockt{\"a}schel}, \bibinfo{person}{Sebastian Riedel}, {and} \bibinfo{person}{Douwe Kiela}.} \bibinfo{year}{2020}\natexlab{}.
\newblock \showarticletitle{Retrieval-Augmented Generation for Knowledge-Intensive {NLP} Tasks}. In \bibinfo{booktitle}{\emph{NeurIPS'20}}. \bibinfo{pages}{9459--9474}.
\newblock


\bibitem[Li et~al\mbox{.}(2026)]%
        {li2026openresearcher}
\bibfield{author}{\bibinfo{person}{Zhuofeng Li}, \bibinfo{person}{Dongfu Jiang}, \bibinfo{person}{Xueguang Ma}, \bibinfo{person}{Haoxiang Zhang}, \bibinfo{person}{Ping Nie}, \bibinfo{person}{Yuyu Zhang}, \bibinfo{person}{Kai Zou}, \bibinfo{person}{Jianwen Xie}, \bibinfo{person}{Yu Zhang}, {and} \bibinfo{person}{Wenhu Chen}.} \bibinfo{year}{2026}\natexlab{}.
\newblock \showarticletitle{{OpenResearcher}: A Fully Open Pipeline for Long-Horizon Deep Research Trajectory Synthesis}. In \bibinfo{booktitle}{\emph{EMNLP'26}}. \bibinfo{pages}{908--937}.
\newblock


\bibitem[Lin(2004)]%
        {lin2004rouge}
\bibfield{author}{\bibinfo{person}{Chin-Yew Lin}.} \bibinfo{year}{2004}\natexlab{}.
\newblock \showarticletitle{{ROUGE}: A Package for Automatic Evaluation of Summaries}. In \bibinfo{booktitle}{\emph{ACL'04 Workshop on Text Summarization}}. \bibinfo{pages}{74--81}.
\newblock


\bibitem[Lin and Och(2004)]%
        {lin2004automatic}
\bibfield{author}{\bibinfo{person}{Chin-Yew Lin} {and} \bibinfo{person}{Franz~Josef Och}.} \bibinfo{year}{2004}\natexlab{}.
\newblock \showarticletitle{Automatic Evaluation of Machine Translation Quality Using Longest Common Subsequence and Skip-Bigram Statistics}. In \bibinfo{booktitle}{\emph{ACL'04}}. \bibinfo{pages}{605--612}.
\newblock


\bibitem[Liu(2025a)]%
        {liu2025grapho1}
\bibfield{author}{\bibinfo{person}{Lihui Liu}.} \bibinfo{year}{2025}\natexlab{a}.
\newblock \showarticletitle{{Graph-O1}: Monte Carlo Tree Search with Reinforcement Learning for Text-Attributed Graph Reasoning}.
\newblock \bibinfo{journal}{\emph{arXiv preprint arXiv:2512.17912}} (\bibinfo{year}{2025}).
\newblock


\bibitem[Liu(2025b)]%
        {liu2025graphmcts}
\bibfield{author}{\bibinfo{person}{Lihui Liu}.} \bibinfo{year}{2025}\natexlab{b}.
\newblock \showarticletitle{Monte Carlo Tree Search for Graph Reasoning in Large Language Model Agents}. In \bibinfo{booktitle}{\emph{CIKM'25}}. \bibinfo{pages}{4966--4970}.
\newblock


\bibitem[Luo et~al\mbox{.}(2026)]%
        {luo2025graph}
\bibfield{author}{\bibinfo{person}{Haoran Luo}, \bibinfo{person}{Haihong E}, \bibinfo{person}{Guanting Chen}, \bibinfo{person}{Qika Lin}, \bibinfo{person}{Yikai Guo}, \bibinfo{person}{Fangzhi Xu}, \bibinfo{person}{Zemin Kuang}, \bibinfo{person}{Meina Song}, \bibinfo{person}{Xiaobao Wu}, \bibinfo{person}{Yifan Zhu}, {and} \bibinfo{person}{Anh~Tuan Luu}.} \bibinfo{year}{2026}\natexlab{}.
\newblock \showarticletitle{{Graph-R1}: Towards Agentic {GraphRAG} Framework via End-to-End Reinforcement Learning}. In \bibinfo{booktitle}{\emph{ICML'26}}.
\newblock


\bibitem[Luo et~al\mbox{.}(2024)]%
        {luo2024reasoning}
\bibfield{author}{\bibinfo{person}{Linhao Luo}, \bibinfo{person}{Yuan-Fang Li}, \bibinfo{person}{Gholamreza Haffari}, {and} \bibinfo{person}{Shirui Pan}.} \bibinfo{year}{2024}\natexlab{}.
\newblock \showarticletitle{Reasoning on Graphs: Faithful and Interpretable Large Language Model Reasoning}. In \bibinfo{booktitle}{\emph{ICLR'24}}.
\newblock


\bibitem[Ma et~al\mbox{.}(2025)]%
        {ma2025think}
\bibfield{author}{\bibinfo{person}{Shengjie Ma}, \bibinfo{person}{Chengjin Xu}, \bibinfo{person}{Xuhui Jiang}, \bibinfo{person}{Muzhi Li}, \bibinfo{person}{Huaren Qu}, \bibinfo{person}{Cehao Yang}, \bibinfo{person}{Jiaxin Mao}, {and} \bibinfo{person}{Jian Guo}.} \bibinfo{year}{2025}\natexlab{}.
\newblock \showarticletitle{{Think-on-Graph} 2.0: Deep and Faithful Large Language Model Reasoning with Knowledge-Guided Retrieval Augmented Generation}. In \bibinfo{booktitle}{\emph{ICLR'25}}.
\newblock


\bibitem[{OpenAI}(2023)]%
        {openai2023gpt35turbo}
\bibfield{author}{\bibinfo{person}{{OpenAI}}.} \bibinfo{year}{2023}\natexlab{}.
\newblock \bibinfo{title}{{GPT-3.5 Turbo}}.
\newblock
\shownote{Accessed: 2026-10-25}.
\newblock
\urldef\tempurl%
\url{https://developers.openai.com/api/docs/models/gpt-3.5-turbo}
\showURL{%
\tempurl}


\bibitem[{OpenAI}(2025a)]%
        {openai2025gpt41mini}
\bibfield{author}{\bibinfo{person}{{OpenAI}}.} \bibinfo{year}{2025}\natexlab{a}.
\newblock \bibinfo{title}{{GPT-4.1 mini}}.
\newblock
\shownote{Accessed: 2026-10-24}.
\newblock
\urldef\tempurl%
\url{https://developers.openai.com/api/docs/models/gpt-4.1-mini}
\showURL{%
\tempurl}


\bibitem[{OpenAI}(2025b)]%
        {openai2025gpt5nano}
\bibfield{author}{\bibinfo{person}{{OpenAI}}.} \bibinfo{year}{2025}\natexlab{b}.
\newblock \bibinfo{title}{{GPT-5 nano}}.
\newblock
\shownote{Accessed: 2026-10-24}.
\newblock
\urldef\tempurl%
\url{https://developers.openai.com/api/docs/models/gpt-5-nano}
\showURL{%
\tempurl}


\bibitem[Ouyang et~al\mbox{.}(2022)]%
        {ouyang2022training}
\bibfield{author}{\bibinfo{person}{Long Ouyang}, \bibinfo{person}{Jeffrey Wu}, \bibinfo{person}{Xu Jiang}, \bibinfo{person}{Diogo Almeida}, \bibinfo{person}{Carroll Wainwright}, \bibinfo{person}{Pamela Mishkin}, \bibinfo{person}{Chong Zhang}, \bibinfo{person}{Sandhini Agarwal}, \bibinfo{person}{Katarina Slama}, \bibinfo{person}{Alex Ray}, {et~al\mbox{.}}} \bibinfo{year}{2022}\natexlab{}.
\newblock \showarticletitle{Training Language Models to Follow Instructions with Human Feedback}. In \bibinfo{booktitle}{\emph{NeurIPS'22}}. \bibinfo{pages}{27730--27744}.
\newblock


\bibitem[Parashar et~al\mbox{.}(2026)]%
        {parashar2025curriculum}
\bibfield{author}{\bibinfo{person}{Shubham Parashar}, \bibinfo{person}{Shurui Gui}, \bibinfo{person}{Xiner Li}, \bibinfo{person}{Hongyi Ling}, \bibinfo{person}{Sushil Vemuri}, \bibinfo{person}{Blake Olson}, \bibinfo{person}{Eric Li}, \bibinfo{person}{Yu Zhang}, \bibinfo{person}{James Caverlee}, \bibinfo{person}{Dileep Kalathil}, {and} \bibinfo{person}{Shuiwang Ji}.} \bibinfo{year}{2026}\natexlab{}.
\newblock \showarticletitle{Curriculum Reinforcement Learning from Easy to Hard Tasks Improves {LLM} Reasoning}. In \bibinfo{booktitle}{\emph{ICLR'26}}.
\newblock


\bibitem[Schulman et~al\mbox{.}(2016)]%
        {schulman2016high}
\bibfield{author}{\bibinfo{person}{John Schulman}, \bibinfo{person}{Philipp Moritz}, \bibinfo{person}{Sergey Levine}, \bibinfo{person}{Michael~I. Jordan}, {and} \bibinfo{person}{Pieter Abbeel}.} \bibinfo{year}{2016}\natexlab{}.
\newblock \showarticletitle{High-Dimensional Continuous Control Using Generalized Advantage Estimation}. In \bibinfo{booktitle}{\emph{ICLR'16}}.
\newblock


\bibitem[Schulman et~al\mbox{.}(2017)]%
        {schulman2017proximal}
\bibfield{author}{\bibinfo{person}{John Schulman}, \bibinfo{person}{Filip Wolski}, \bibinfo{person}{Prafulla Dhariwal}, \bibinfo{person}{Alec Radford}, {and} \bibinfo{person}{Oleg Klimov}.} \bibinfo{year}{2017}\natexlab{}.
\newblock \showarticletitle{Proximal Policy Optimization Algorithms}.
\newblock \bibinfo{journal}{\emph{arXiv preprint arXiv:1707.06347}} (\bibinfo{year}{2017}).
\newblock


\bibitem[Shao et~al\mbox{.}(2024)]%
        {shao2024deepseekmath}
\bibfield{author}{\bibinfo{person}{Zhihong Shao}, \bibinfo{person}{Peiyi Wang}, \bibinfo{person}{Qihao Zhu}, \bibinfo{person}{Runxin Xu}, \bibinfo{person}{Junxiao Song}, \bibinfo{person}{Xiao Bi}, \bibinfo{person}{Haowei Zhang}, \bibinfo{person}{Mingchuan Zhang}, \bibinfo{person}{Y.~K. Li}, \bibinfo{person}{Y. Wu}, {and} \bibinfo{person}{Daya Guo}.} \bibinfo{year}{2024}\natexlab{}.
\newblock \showarticletitle{{DeepSeekMath}: Pushing the Limits of Mathematical Reasoning in Open Language Models}.
\newblock \bibinfo{journal}{\emph{arXiv preprint arXiv:2402.03300}} (\bibinfo{year}{2024}).
\newblock


\bibitem[Shen et~al\mbox{.}(2025)]%
        {shen2025thinking}
\bibfield{author}{\bibinfo{person}{Junhong Shen}, \bibinfo{person}{Hao Bai}, \bibinfo{person}{Lunjun Zhang}, \bibinfo{person}{Yifei Zhou}, \bibinfo{person}{Amrith Setlur}, \bibinfo{person}{Peter Tong}, \bibinfo{person}{Diego Caples}, \bibinfo{person}{Nan Jiang}, \bibinfo{person}{Tong Zhang}, \bibinfo{person}{Ameet~S. Talwalkar}, {and} \bibinfo{person}{Aviral Kumar}.} \bibinfo{year}{2025}\natexlab{}.
\newblock \showarticletitle{Thinking vs. Doing: Improving Agent Reasoning by Scaling Test-Time Interaction}. In \bibinfo{booktitle}{\emph{NeurIPS'25}}. \bibinfo{pages}{187840--187881}.
\newblock


\bibitem[Singhal et~al\mbox{.}(2025)]%
        {bercovich2025llama}
\bibfield{author}{\bibinfo{person}{Soumye Singhal}, \bibinfo{person}{Jiaqi Zeng}, \bibinfo{person}{Alexander Bukharin}, \bibinfo{person}{Yian Zhang}, \bibinfo{person}{Gerald Shen}, \bibinfo{person}{Ameya~Sunil Mahabaleshwarkar}, \bibinfo{person}{Bilal Kartal}, \bibinfo{person}{Yoshi Suhara}, \bibinfo{person}{Akhiad Bercovich}, \bibinfo{person}{Itay Levy}, {et~al\mbox{.}}} \bibinfo{year}{2025}\natexlab{}.
\newblock \showarticletitle{{Llama-Nemotron}: Efficient Reasoning Models}. In \bibinfo{booktitle}{\emph{ICML'25 Workshop on EXAIT}}.
\newblock


\bibitem[Song et~al\mbox{.}(2025)]%
        {song2025r1searcher}
\bibfield{author}{\bibinfo{person}{Huatong Song}, \bibinfo{person}{Jinhao Jiang}, \bibinfo{person}{Yingqian Min}, \bibinfo{person}{Jie Chen}, \bibinfo{person}{Zhipeng Chen}, \bibinfo{person}{Wayne~Xin Zhao}, \bibinfo{person}{Lei Fang}, {and} \bibinfo{person}{Ji-Rong Wen}.} \bibinfo{year}{2025}\natexlab{}.
\newblock \showarticletitle{{R1-Searcher}: Incentivizing the Search Capability in {LLMs} via Reinforcement Learning}.
\newblock \bibinfo{journal}{\emph{arXiv preprint arXiv:2503.05592}} (\bibinfo{year}{2025}).
\newblock


\bibitem[Sun et~al\mbox{.}(2024a)]%
        {sun2024think}
\bibfield{author}{\bibinfo{person}{Jiashuo Sun}, \bibinfo{person}{Chengjin Xu}, \bibinfo{person}{Lumingyuan Tang}, \bibinfo{person}{Saizhuo Wang}, \bibinfo{person}{Chen Lin}, \bibinfo{person}{Yeyun Gong}, \bibinfo{person}{Lionel~M. Ni}, \bibinfo{person}{Heung-Yeung Shum}, {and} \bibinfo{person}{Jian Guo}.} \bibinfo{year}{2024}\natexlab{a}.
\newblock \showarticletitle{{Think-on-Graph}: Deep and Responsible Reasoning of Large Language Model on Knowledge Graph}. In \bibinfo{booktitle}{\emph{ICLR'24}}.
\newblock


\bibitem[Sun et~al\mbox{.}(2024b)]%
        {sun2024head}
\bibfield{author}{\bibinfo{person}{Kai Sun}, \bibinfo{person}{Yifan Xu}, \bibinfo{person}{Hanwen Zha}, \bibinfo{person}{Yue Liu}, {and} \bibinfo{person}{Xin~Luna Dong}.} \bibinfo{year}{2024}\natexlab{b}.
\newblock \showarticletitle{{Head-to-Tail}: How Knowledgeable Are Large Language Models ({LLMs})? {A.K.A.} Will {LLMs} Replace Knowledge Graphs\bibqmark}. In \bibinfo{booktitle}{\emph{NAACL'24}}. \bibinfo{pages}{311--325}.
\newblock


\bibitem[Vu et~al\mbox{.}(2024)]%
        {vu2024freshllms}
\bibfield{author}{\bibinfo{person}{Tu Vu}, \bibinfo{person}{Mohit Iyyer}, \bibinfo{person}{Xuezhi Wang}, \bibinfo{person}{Noah Constant}, \bibinfo{person}{Jerry Wei}, \bibinfo{person}{Jason Wei}, \bibinfo{person}{Chris Tar}, \bibinfo{person}{Yun-Hsuan Sung}, \bibinfo{person}{Denny Zhou}, \bibinfo{person}{Quoc Le}, {and} \bibinfo{person}{Thang Luong}.} \bibinfo{year}{2024}\natexlab{}.
\newblock \showarticletitle{{FreshLLMs}: Refreshing Large Language Models with Search Engine Augmentation}. In \bibinfo{booktitle}{\emph{Findings of ACL'24}}. \bibinfo{pages}{13697--13720}.
\newblock


\bibitem[Wu et~al\mbox{.}(2025a)]%
        {wu2025webdancer}
\bibfield{author}{\bibinfo{person}{Jialong Wu}, \bibinfo{person}{Baixuan Li}, \bibinfo{person}{Runnan Fang}, \bibinfo{person}{Wenbiao Yin}, \bibinfo{person}{Liwen Zhang}, \bibinfo{person}{Zhenglin Wang}, \bibinfo{person}{Zhengwei Tao}, \bibinfo{person}{Ding-Chu Zhang}, \bibinfo{person}{Zekun Xi}, \bibinfo{person}{Robert Tang}, \bibinfo{person}{Yong Jiang}, \bibinfo{person}{Pengjun Xie}, \bibinfo{person}{Fei Huang}, {and} \bibinfo{person}{Jingren Zhou}.} \bibinfo{year}{2025}\natexlab{a}.
\newblock \showarticletitle{{WebDancer}: Towards Autonomous Information Seeking Agency}. In \bibinfo{booktitle}{\emph{NeurIPS'25}}.
\newblock


\bibitem[Wu et~al\mbox{.}(2025b)]%
        {wu2025structure}
\bibfield{author}{\bibinfo{person}{Junlin Wu}, \bibinfo{person}{Xianrui Zhong}, \bibinfo{person}{Jiashuo Sun}, \bibinfo{person}{Bolian Li}, \bibinfo{person}{Bowen Jin}, \bibinfo{person}{Jiawei Han}, {and} \bibinfo{person}{Qingkai Zeng}.} \bibinfo{year}{2025}\natexlab{b}.
\newblock \showarticletitle{{Structure-R1}: Dynamically Leveraging Structural Knowledge in {LLM} Reasoning through Reinforcement Learning}.
\newblock \bibinfo{journal}{\emph{arXiv preprint arXiv:2510.15191}} (\bibinfo{year}{2025}).
\newblock


\bibitem[Yang et~al\mbox{.}(2025)]%
        {yang2025qwen3}
\bibfield{author}{\bibinfo{person}{An Yang}, \bibinfo{person}{Anfeng Li}, \bibinfo{person}{Baosong Yang}, \bibinfo{person}{Beichen Zhang}, \bibinfo{person}{Binyuan Hui}, \bibinfo{person}{Bo Zheng}, \bibinfo{person}{Bowen Yu}, \bibinfo{person}{Chang Gao}, \bibinfo{person}{Chengen Huang}, \bibinfo{person}{Chenxu Lv}, {et~al\mbox{.}}} \bibinfo{year}{2025}\natexlab{}.
\newblock \showarticletitle{{Qwen3} Technical Report}.
\newblock \bibinfo{journal}{\emph{arXiv preprint arXiv:2505.09388}} (\bibinfo{year}{2025}).
\newblock


\bibitem[Yang et~al\mbox{.}(2024)]%
        {hui2024qwen2}
\bibfield{author}{\bibinfo{person}{An Yang}, \bibinfo{person}{Baosong Yang}, \bibinfo{person}{Beichen Zhang}, \bibinfo{person}{Binyuan Hui}, \bibinfo{person}{Bo Zheng}, \bibinfo{person}{Bowen Yu}, \bibinfo{person}{Chengyuan Li}, \bibinfo{person}{Dayiheng Liu}, \bibinfo{person}{Fei Huang}, \bibinfo{person}{Haoran Wei}, {et~al\mbox{.}}} \bibinfo{year}{2024}\natexlab{}.
\newblock \showarticletitle{{Qwen2.5} Technical Report}.
\newblock \bibinfo{journal}{\emph{arXiv preprint arXiv:2412.15115}} (\bibinfo{year}{2024}).
\newblock


\bibitem[Ye et~al\mbox{.}(2020)]%
        {ye2020mastering}
\bibfield{author}{\bibinfo{person}{Deheng Ye}, \bibinfo{person}{Zhao Liu}, \bibinfo{person}{Mingfei Sun}, \bibinfo{person}{Bei Shi}, \bibinfo{person}{Peilin Zhao}, \bibinfo{person}{Hao Wu}, \bibinfo{person}{Hongsheng Yu}, \bibinfo{person}{Shaojie Yang}, \bibinfo{person}{Xipeng Wu}, \bibinfo{person}{Qingwei Guo}, {et~al\mbox{.}}} \bibinfo{year}{2020}\natexlab{}.
\newblock \showarticletitle{Mastering Complex Control in {MOBA} Games with Deep Reinforcement Learning}. In \bibinfo{booktitle}{\emph{AAAI'20}}. \bibinfo{pages}{6672--6679}.
\newblock


\bibitem[Ying et~al\mbox{.}(2026)]%
        {ying2026graphscout}
\bibfield{author}{\bibinfo{person}{Yuchen Ying}, \bibinfo{person}{Weiqi Jiang}, \bibinfo{person}{Tongya Zheng}, \bibinfo{person}{Yu Wang}, \bibinfo{person}{Shunyu Liu}, \bibinfo{person}{Kaixuan Chen}, {and} \bibinfo{person}{Mingli Song}.} \bibinfo{year}{2026}\natexlab{}.
\newblock \showarticletitle{{GraphScout}: Empowering Large Language Models with Intrinsic Exploration Ability for Agentic Graph Reasoning}. In \bibinfo{booktitle}{\emph{KDD'26}}. \bibinfo{pages}{6258--6269}.
\newblock


\end{thebibliography}

\appendix
\section{Implementation Details}
\label{app:implementation}

\subsection{Training Procedure and Optimization}
\label{app:optimization}

Algorithm~\ref{alg:training} summarizes the training procedure.
Table~\ref{tab:hyperparameters} gives the shared settings and backbone
differences. The actor and critic start from the corresponding original
backbone, with a value head added to the critic.

\begin{table}[htbp]
\centering
\small
\caption{Training and rollout settings (Qwen2.5-3B: Qwen2.5-3B-Instruct).
Each context limit caps the tokens of its component separately.}
\label{tab:hyperparameters}
\begin{tabular*}{\linewidth}{@{\extracolsep{\fill}}lcc@{}}
\toprule
Setting & Qwen2.5-3B & Qwen3-14B \\
\midrule
Training / validation / in-domain test & \multicolumn{2}{c}{749 / 107 / 214} \\
Unseen-domain test questions & \multicolumn{2}{c}{620} \\
Training updates & \multicolumn{2}{c}{200} \\
Trajectories / minibatch & \multicolumn{2}{c}{128 / 128} \\
Rollouts per question / epochs & \multicolumn{2}{c}{1 / 1} \\
Actor / critic learning rate & \multicolumn{2}{c}{$10^{-6}$ / $10^{-5}$} \\
Actor / critic warmup updates & \multicolumn{2}{c}{56 / 3} \\
Learning rate after warmup & \multicolumn{2}{c}{Constant} \\
GAE $(\gamma,\lambda)$ & \multicolumn{2}{c}{$(1,1)$} \\
Policy / value clip range & \multicolumn{2}{c}{$0.2$ / $0.5$} \\
Dual clip / gradient clip & \multicolumn{2}{c}{$3$ / $1$} \\
Entropy coefficient & \multicolumn{2}{c}{$0.001$} \\
Reference KL in reward / loss & \multicolumn{2}{c}{Off / off} \\
Micro-batch per GPU & 8 & 1 \\
Agent-token loss mean & \multicolumn{2}{c}{Per micro-batch} \\
\arrayrulecolor{gdRule}\midrule\arrayrulecolor{black}
Reward weights $(\alpha_C,\alpha_F,\alpha_H)$ & \multicolumn{2}{c}{$(0.8,0.2,0.1)$} \\
Pacing $\mu(u)$ / kernel width $\sigma$ & \multicolumn{2}{c}{$2u^3$ / $0.75$} \\
Mixing schedule $\eta(u)$ & \multicolumn{2}{c}{$0.2+0.6u$} \\
Coverage prior $q$ & \multicolumn{2}{c}{$(0.34,0.33,0.33)$} \\
\arrayrulecolor{gdRule}\midrule\arrayrulecolor{black}
Temperature / top-$p$ / top-$k$ & \multicolumn{2}{c}{$0.7$ / $0.8$ / $20$} \\
New tokens per generation & \multicolumn{2}{c}{500} \\
Observation tokens & \multicolumn{2}{c}{1,024} \\
Rolling context: train / eval & \multicolumn{2}{c}{24,576 / 30,000} \\
Initial prompt retained for PPO & 1,024 & 8,192 \\
Tool rounds + final generation & \multicolumn{2}{c}{10 + 1} \\
Observation continuation & None & \texttt{<think>} \\
\arrayrulecolor{gdRule}\midrule\arrayrulecolor{black}
Training H200 GPUs & 4 & 16 \\
Tensor / sequence parallel size & 1 / 1 & 2 / 2 \\
\bottomrule
\end{tabular*}
\end{table}

\begin{table*}[tp]
\centering
\small
\caption{Shared instruction template, with domain-specific content shown
as placeholders.}
\label{tab:prompt_template}
\begin{tabularx}{\linewidth}{@{}p{0.115\linewidth}X@{}}
\toprule
Task & \texttt{\detokenize{Solve a question answering task by repeating bundled steps that contain reasoning (<think>...</think>) followed by exactly one graph interaction (<graph>...</graph>). After each <graph> call, the environment returns feedback inside <information>...</information>. You may take as many steps as necessary.}} \\
\arrayrulecolor{gdRule}\midrule\arrayrulecolor{black}
Protocol & \texttt{\detokenize{Output protocol (bundled steps):}}\newline
\texttt{\detokenize{- Intermediate step (must include BOTH in a single output, in this order):}}\newline
\texttt{\detokenize{<think>...</think><graph>Function[...]</graph>}}\newline
\texttt{\detokenize{(Then the environment responds with <information>...</information>.)}}\newline
\texttt{\detokenize{- Final step (no more graph calls):}}\newline
\texttt{\detokenize{<think>...</think><answer>...</answer>}} \\
\arrayrulecolor{gdRule}\midrule\arrayrulecolor{black}
Functions & \texttt{\detokenize{Available graph interaction functions (use exactly these names and signatures INSIDE <graph>...):}}\newline
\texttt{\detokenize{- RetrieveNode[keyword]} \# \detokenize{retrieves the related node from the graph according to the query}}\newline
\texttt{\detokenize{- NodeFeature[Node, feature]} \# \detokenize{returns detailed attribute information of Node for the given "feature" key}}\newline
\texttt{\detokenize{- NodeDegree[Node, neighbor_type]} \# \detokenize{returns the number of "neighbor_type" neighbors of Node}}\newline
\texttt{\detokenize{- NeighbourCheck[Node, neighbor_type]} \# \detokenize{lists the "neighbor_type" neighbors of Node and returns them}} \\
\arrayrulecolor{gdRule}\midrule\arrayrulecolor{black}
Rules & \texttt{\detokenize{Rules and format:}}\newline
\texttt{\detokenize{1) You MUST conduct reasoning inside <think>...</think> before every graph call and after every <information> you receive.}}\newline
\texttt{\detokenize{2) Inside <graph>...</graph>, issue EXACTLY ONE function from the list below per step. Do NOT include any other text in <graph>.}}\newline
\texttt{\detokenize{3) Do NOT fabricate <information>; it is ONLY produced by the environment immediately after your <graph> step.}}\newline
\texttt{\detokenize{4) Keep thoughts concise and ONLY inside <think>. Do NOT put a graph call inside <think>, and do NOT put thoughts inside <graph>.}}\newline
\texttt{\detokenize{5) The final output MUST contain ONLY one <answer>...</answer> block with the requested node main features (e.g., names), not node IDs.}} \\
\arrayrulecolor{gdRule}\midrule\arrayrulecolor{black}
Examples & \texttt{\detokenize{Here are some examples:}}\newline
\textcolor{gdBlue}{\texttt{\{examples\}}}\newline
\texttt{\detokenize{(END OF EXAMPLES)}} \\
\arrayrulecolor{gdRule}\midrule\arrayrulecolor{black}
Graph schema & \texttt{\detokenize{Definition of the graph:}}\newline
\textcolor{gdBlue}{\texttt{\{graph\_definition\}}} \\
\arrayrulecolor{gdRule}\midrule\arrayrulecolor{black}
Question & \texttt{\detokenize{Question:}}\newline
\textcolor{gdBlue}{\texttt{\{question\}}} \\
\bottomrule
\end{tabularx}
\end{table*}

For a return target $V^{\rm targ}_i$ and the pre-update value $V_{\rm old}$,
the critic minimizes
\begin{equation}
 \tfrac12\mathbb E_i\!\left[
 \max\left\{(V_\phi(s_i)-V^{\rm targ}_i)^2,\,
 (\widetilde V_\phi(s_i)-V^{\rm targ}_i)^2\right\}\right],
\end{equation}
where $\widetilde V_\phi=\operatorname{clip}(V_\phi,V_{\rm old}-0.5,
V_{\rm old}+0.5)$. Only agent-generated positions contribute.
Returns are formed before masked advantage normalization. Reward KL and
the actor reference-KL loss are disabled.

Reward exact match lowercases text, removes ASCII punctuation and the
English articles \emph{a}, \emph{an}, and \emph{the}, and collapses whitespace.
An answer counts as correct if it equals the normalized reference.
The format validator checks the order and closure of reasoning, action,
observation, and answer tags and rejects text outside these blocks. It
ignores node IDs and relations.

\subsection{Interaction Protocol and Budgets}
\label{app:protocol}
\label{app:prompt}

{\raggedright
Table~\ref{tab:prompt_template} lists four executable functions:
\texttt{RetrieveNode[keyword]},
\texttt{NodeFeature[Node, feature]},
\texttt{NodeDegree[Node, neighbor\_type]}, and
\texttt{NeighbourCheck[Node, neighbor\_type]}.
The executor returns observations between model outputs and stops further
graph execution after a final answer.\par}

In the rollout prompt (Table~\ref{tab:prompt_template}), the placeholders expand to the domain's
demonstrations, graph schema, and question. The resulting text forms one
user message, wrapped in the backbone's chat template. For Qwen3, the
initial template enables thinking.

The 500-token generation budget covers one round, including reasoning
and its action or answer. The 1,024-token observation budget and the rolling
context window apply separately. After an observation, the executor supplies
Qwen3's opening \texttt{<think>} as context, which counts toward the
observation budget and is masked from optimization. Qwen2.5 uses the same
explicit interaction instructions, and only Qwen3 receives the prefix.
After ten rounds, the rollout allows a final generation with tool
execution disabled.

\subsection{Partition and Structural Labels}
\label{app:labels}

Table~\ref{tab:data_stats} gives the group sizes.
The 1,070-question \textsc{Academic}/\allowbreak\textsc{Biomedical} pool excludes the 50 \textsc{Academic}
questions of the OOD question type. Within each graph, the fixed
partition allocates 70\% to training, 10\% to validation, and 20\% to
in-domain testing, with question-template stratification. \textsc{Academic}
contributes 560/80/160 training/validation/test questions and
\textsc{Biomedical} 189/27/54.
Gradient updates use only the training questions.

The labeler reads $O(x)$ in Eq.~\eqref{eq:difficulty} from template cues.
Aggregation, superlative, or ranking cues mark set-level selection, and
multiple entity slots or additional constraints mark composition. Every
matched template includes single-anchor access, so the maximum assigns
Hard, then Medium, then Easy in this priority order. Counts that one
\texttt{NodeDegree} call answers fall under single-anchor access.
Unmatched templates default to Medium.
Template IDs are namespaced by graph family. The curriculum and the in-domain difficulty
analyses use these labels. The OOD question-type subset is defined separately from both
this three-level assignment and our unseen-domain definition.

The 550 unseen-domain questions outside the OOD group are labeled by LLM-assisted
classification (\texttt{gpt-6-luna}) followed by template-family review
against the graph schema and an operation-based rubric.
The 70 OOD questions form a disjoint group.

\subsection{Evaluation and Baseline Settings}

ROUGE-L for \ours, Vanilla PPO, and Graph-CoT with Qwen backbones uses
unstemmed tokenization in Tables~\ref{tab:main}--\ref{tab:difficulty_results}
and~\ref{tab:main_3b} and Figure~\ref{fig:avg_scores}.
Graph-CoT with
GPT-4.1-mini (\texttt{gpt-\allowbreak 4.1-mini-\allowbreak 2025-04-14}) uses the official agent code and few-shot examples. Its instruction extends
the original one with an explicit Finish action and strict Thought/Action
alternation. Each step makes a Thought call and an Action call at
temperature 0 with at most 300 output tokens each, for up to ten steps. Graph Counselor with GPT-4.1-mini uses
acting and reflection temperature 0.7, judge temperature 0.3, ten steps,
and up to two reflections. Its acting, reflection, and judging calls have
output limits of 512, 1,024, and 256 tokens, respectively.
GPT-5-nano (\texttt{gpt-\allowbreak 5-nano-\allowbreak 2025-08-07}) calls use the model's
default sampling, medium reasoning effort, and a 25,000-token completion
budget.

The Graph Counselor results in \ifsubmission Table~\ref{tab:eh}\else
Tables~\ref{tab:eh}, \ref{tab:domain_behavior},
and~\ref{tab:difficulty_behavior}\fi{} come from separate diagnostic runs
that log tool observations.

\subsection{Use of AI Assistants}
We used AI assistants for coding support and for polishing the writing. The
authors reviewed all code and text and checked every reported number against
its raw output or cited source.

\section{Additional Experimental Analysis}
\label{app:analysis}

Table~\ref{tab:main_3b} gives the per-column results with
Qwen2.5-3B-Instruct that Figure~\ref{fig:avg_scores} summarizes.

\begin{table*}[tbp]
\centering
\small
\setlength{\tabcolsep}{5.2pt}
\caption{Results with Qwen2.5-3B-Instruct. Both baselines run the original
backbone. Columns follow Table~\ref{tab:main}. Bold: best per column.
{\boldmath$\Delta$\%: per-column gain $100\,(s^{\ast}-s)/s$ of \ours{} (3B, $s^{\ast}$) over the row ($s$),
averaged over the four domain columns, for R-L and GS separately.}}
\label{tab:main_3b}
\begin{tabular}{lrrrrrrrrrrrr}
\toprule
 & \multicolumn{2}{c}{In-domain} & \multicolumn{2}{c}{\textsc{E-commerce}} & \multicolumn{2}{c}{\textsc{Literature}} & \multicolumn{2}{c}{\textsc{Legal}} & \multicolumn{2}{c}{AVG} & \multicolumn{2}{c}{$\Delta$\%} \\
\cmidrule(lr){2-3}\cmidrule(lr){4-5}\cmidrule(lr){6-7}\cmidrule(lr){8-9}\cmidrule(lr){10-11}\cmidrule(lr){12-13}
Method & R-L & GS & R-L & GS & R-L & GS & R-L & GS & R-L & GS & \multicolumn{1}{c}{R-L} & \multicolumn{1}{c}{GS} \\
\midrule
Graph-CoT~\cite{jin2024graph} & 29.36 & 32.24 & 43.67 & 45.00 & 40.00 & 45.42 & 31.85 & 34.44 & 36.22 & 39.28 & \textcolor{gdGreen}{$+26.5$} & \textcolor{gdGreen}{$+25.2$} \\
Graph Counselor~\cite{gao2025graph} & 27.77 & 28.50 & 42.19 & 41.00 & 41.93 & 48.33 & 15.06 & 22.78 & 31.74 & 35.15 & \textcolor{gdGreen}{$+62.2$} & \textcolor{gdGreen}{$+47.3$} \\
\rowcolor{gdBlueTint}
\ours & \textbf{45.65} & \textbf{45.33} & \textbf{48.92} & \textbf{50.00} & \textbf{46.89} & \textbf{54.58} & \textbf{38.62} & \textbf{44.44} & \textbf{45.02} & \textbf{48.59} & \textcolor{gdGray}{--} & \textcolor{gdGray}{--} \\
\bottomrule
\end{tabular}
\end{table*}

\ifsubmission\else
\begin{table}[tbp]
\centering
\small
\setlength{\tabcolsep}{4pt}
\caption{Behavioral analysis (\%). VF (valid format) measures
complete protocol conformity over questions, VC (valid calls) basic syntax
validity over proposed calls, and EH lexical evidence hit over
questions. -- denotes an inapplicable metric.}
\label{tab:domain_behavior}
\begin{tabularx}{\linewidth}{@{}ll*{3}{>{\raggedleft\arraybackslash}X}@{}}
\toprule
Domain & Method & VF & VC & EH \\
\midrule
 & Graph-CoT (3B) & 32.24 & 97.59 & 39.72 \\
 & Graph Counselor (3B) & \textcolor{gdGray}{--} & \textcolor{gdGray}{--} & 47.66 \\
 & Vanilla PPO (3B) & 49.53 & 95.02 & 48.60 \\
\rowcolor{gdBlueTint}\cellcolor{white} & \ours{} (3B) & 86.92 & 96.89 & \multicolumn{1}{>{\columncolor{gdBlueTint}[\tabcolsep][0pt]}r@{}}{53.27} \\
 & Graph-CoT (14B) & 32.71 & 99.74 & 61.68 \\
\rowcolor{gdBlueTint}\cellcolor{white}\multirow{-6}{*}{In-domain} & \ours{} (14B) & 67.29 & 100.00 & \multicolumn{1}{>{\columncolor{gdBlueTint}[\tabcolsep][0pt]}r@{}}{66.82} \\
\arrayrulecolor{gdRule}\midrule\arrayrulecolor{black}
 & Graph-CoT (3B) & 60.00 & 99.76 & 50.50 \\
 & Graph Counselor (3B) & \textcolor{gdGray}{--} & \textcolor{gdGray}{--} & 50.50 \\
 & Vanilla PPO (3B) & 69.00 & 96.87 & 47.50 \\
\rowcolor{gdBlueTint}\cellcolor{white} & \ours{} (3B) & 90.50 & 94.81 & \multicolumn{1}{>{\columncolor{gdBlueTint}[\tabcolsep][0pt]}r@{}}{52.00} \\
 & Graph-CoT (14B) & 30.50 & 99.32 & 56.00 \\
\rowcolor{gdBlueTint}\cellcolor{white}\multirow{-6}{*}{\textsc{E-commerce}} & \ours{} (14B) & 62.50 & 100.00 & \multicolumn{1}{>{\columncolor{gdBlueTint}[\tabcolsep][0pt]}r@{}}{56.00} \\
\arrayrulecolor{gdRule}\midrule\arrayrulecolor{black}
 & Graph-CoT (3B) & 46.25 & 100.00 & 51.25 \\
 & Graph Counselor (3B) & \textcolor{gdGray}{--} & \textcolor{gdGray}{--} & 56.25 \\
 & Vanilla PPO (3B) & 67.08 & 92.85 & 54.17 \\
\rowcolor{gdBlueTint}\cellcolor{white} & \ours{} (3B) & 88.33 & 95.87 & \multicolumn{1}{>{\columncolor{gdBlueTint}[\tabcolsep][0pt]}r@{}}{55.42} \\
 & Graph-CoT (14B) & 39.17 & 99.16 & 62.08 \\
\rowcolor{gdBlueTint}\cellcolor{white}\multirow{-6}{*}{\textsc{Literature}} & \ours{} (14B) & 70.00 & 100.00 & \multicolumn{1}{>{\columncolor{gdBlueTint}[\tabcolsep][0pt]}r@{}}{63.33} \\
\arrayrulecolor{gdRule}\midrule\arrayrulecolor{black}
 & Graph-CoT (3B) & 47.22 & 96.79 & 46.67 \\
 & Graph Counselor (3B) & \textcolor{gdGray}{--} & \textcolor{gdGray}{--} & 37.78 \\
 & Vanilla PPO (3B) & 70.00 & 90.76 & 44.44 \\
\rowcolor{gdBlueTint}\cellcolor{white} & \ours{} (3B) & 77.78 & 94.28 & \multicolumn{1}{>{\columncolor{gdBlueTint}[\tabcolsep][0pt]}r@{}}{52.22} \\
 & Graph-CoT (14B) & 27.78 & 97.16 & 61.11 \\
\rowcolor{gdBlueTint}\cellcolor{white}\multirow{-6}{*}{\textsc{Legal}} & \ours{} (14B) & 65.56 & 98.66 & \multicolumn{1}{>{\columncolor{gdBlueTint}[\tabcolsep][0pt]}r@{}}{70.00} \\
\bottomrule
\end{tabularx}
\end{table}

\begin{table*}[!tp]
\centering
\small
\renewcommand{\arraystretch}{0.95}
\setlength{\tabcolsep}{3pt}
\caption{Behavior by analysis group (\%). VF, VC, and EH follow
Table~\ref{tab:domain_behavior}, and Table~\ref{tab:data_stats} gives
counts. -- marks an inapplicable protocol metric or an empty subset.
The four panels are mutually exclusive. OOD retains the original
question-type subset.}
\label{tab:difficulty_behavior}
\begin{tabularx}{\linewidth}{@{}l*{12}{>{\raggedleft\arraybackslash}X}@{}}
\toprule
 & \multicolumn{3}{c}{In-domain} & \multicolumn{3}{c}{\textsc{E-commerce}}
 & \multicolumn{3}{c}{\textsc{Literature}} & \multicolumn{3}{c}{\textsc{Legal}} \\
\cmidrule(lr){2-4}\cmidrule(lr){5-7}\cmidrule(lr){8-10}\cmidrule(lr){11-13}
Method & VF & VC & EH & VF & VC & EH & VF & VC & EH & VF & VC & EH \\
\midrule
\multicolumn{13}{@{}l}{\textcolor{gdGray}{\textit{Easy}}} \\
Graph-CoT (3B) & 57.58 & 100.00 & 48.48 & 90.00 & 100.00 & 87.50 & 69.23 & 100.00 & 70.77 & 68.75 & 96.03 & 71.25 \\
Graph Counselor (3B) & \textcolor{gdGray}{--} & \textcolor{gdGray}{--} & 53.03 & \textcolor{gdGray}{--} & \textcolor{gdGray}{--} & 82.50 & \textcolor{gdGray}{--} & \textcolor{gdGray}{--} & 77.69 & \textcolor{gdGray}{--} & \textcolor{gdGray}{--} & 53.75 \\
Vanilla PPO (3B) & 72.73 & 99.58 & 56.06 & 88.75 & 99.50 & 87.50 & 90.00 & 100.00 & 72.31 & 78.75 & 89.02 & 71.25 \\
\rowcolor{gdBlueTint}\multicolumn{1}{@{}>{\columncolor{gdBlueTint}[0pt][\tabcolsep]}l}{\ours{} (3B)} & 95.45 & 98.01 & 56.06 & 100.00 & 99.45 & 93.75 & 94.62 & 98.14 & 75.38 & 90.00 & 91.30 & \multicolumn{1}{>{\columncolor{gdBlueTint}[\tabcolsep][0pt]}r@{}}{77.50} \\
Graph-CoT (14B) & 45.45 & 99.03 & 59.09 & 43.75 & 99.46 & 93.75 & 56.15 & 97.40 & 83.08 & 35.00 & 93.02 & 83.75 \\
\rowcolor{gdBlueTint}\multicolumn{1}{@{}>{\columncolor{gdBlueTint}[0pt][\tabcolsep]}l}{\ours{} (14B)} & 93.94 & 100.00 & 66.67 & 93.75 & 100.00 & 92.50 & 92.31 & 100.00 & 81.54 & 90.00 & 98.37 & \multicolumn{1}{>{\columncolor{gdBlueTint}[\tabcolsep][0pt]}r@{}}{93.75} \\
\arrayrulecolor{gdRule}\midrule\arrayrulecolor{black}
\multicolumn{13}{@{}l}{\textcolor{gdGray}{\textit{Medium}}} \\
Graph-CoT (3B) & 29.87 & 95.54 & 48.05 & 20.00 & 100.00 & 0.00 & 6.67 & 100.00 & 3.33 & 28.00 & 96.84 & 16.00 \\
Graph Counselor (3B) & \textcolor{gdGray}{--} & \textcolor{gdGray}{--} & 59.74 & \textcolor{gdGray}{--} & \textcolor{gdGray}{--} & 0.00 & \textcolor{gdGray}{--} & \textcolor{gdGray}{--} & 0.00 & \textcolor{gdGray}{--} & \textcolor{gdGray}{--} & 10.00 \\
Vanilla PPO (3B) & 45.45 & 89.49 & 55.84 & 10.00 & 100.00 & 0.00 & 20.00 & 88.38 & 0.00 & 66.00 & 89.33 & 10.00 \\
\rowcolor{gdBlueTint}\multicolumn{1}{@{}>{\columncolor{gdBlueTint}[0pt][\tabcolsep]}l}{\ours{} (3B)} & 79.22 & 96.92 & 64.94 & 60.00 & 100.00 & 0.00 & 70.00 & 98.57 & 0.00 & 54.00 & 93.98 & \multicolumn{1}{>{\columncolor{gdBlueTint}[\tabcolsep][0pt]}r@{}}{26.00} \\
Graph-CoT (14B) & 32.47 & 99.77 & 75.32 & 0.00 & 100.00 & 0.00 & 10.00 & 100.00 & 6.67 & 12.00 & 97.84 & 44.00 \\
\rowcolor{gdBlueTint}\multicolumn{1}{@{}>{\columncolor{gdBlueTint}[0pt][\tabcolsep]}l}{\ours{} (14B)} & 72.73 & 100.00 & 80.52 & 0.00 & 100.00 & 0.00 & 33.33 & 100.00 & 16.67 & 36.00 & 99.51 & \multicolumn{1}{>{\columncolor{gdBlueTint}[\tabcolsep][0pt]}r@{}}{58.00} \\
\arrayrulecolor{gdRule}\midrule\arrayrulecolor{black}
\multicolumn{13}{@{}l}{\textcolor{gdGray}{\textit{Hard}}} \\
Graph-CoT (3B) & 11.27 & 98.40 & 22.54 & 40.00 & 99.50 & 44.29 & 22.86 & 100.00 & 42.86 & 26.67 & 96.82 & 33.33 \\
Graph Counselor (3B) & \textcolor{gdGray}{--} & \textcolor{gdGray}{--} & 29.58 & \textcolor{gdGray}{--} & \textcolor{gdGray}{--} & 50.00 & \textcolor{gdGray}{--} & \textcolor{gdGray}{--} & 48.57 & \textcolor{gdGray}{--} & \textcolor{gdGray}{--} & 36.67 \\
Vanilla PPO (3B) & 32.39 & 98.42 & 33.80 & 54.29 & 94.27 & 35.71 & 51.43 & 88.77 & 51.43 & 46.67 & 93.66 & 30.00 \\
\rowcolor{gdBlueTint}\multicolumn{1}{@{}>{\columncolor{gdBlueTint}[0pt][\tabcolsep]}l}{\ours{} (3B)} & 87.32 & 96.18 & 38.03 & 81.43 & 89.66 & 41.43 & 84.29 & 91.38 & 50.00 & 76.67 & 95.98 & \multicolumn{1}{>{\columncolor{gdBlueTint}[\tabcolsep][0pt]}r@{}}{33.33} \\
Graph-CoT (14B) & 21.13 & 100.00 & 49.30 & 18.57 & 100.00 & 52.86 & 25.71 & 100.00 & 55.71 & 23.33 & 99.39 & 40.00 \\
\rowcolor{gdBlueTint}\multicolumn{1}{@{}>{\columncolor{gdBlueTint}[0pt][\tabcolsep]}l}{\ours{} (14B)} & 36.62 & 100.00 & 52.11 & 44.29 & 100.00 & 54.29 & 52.86 & 100.00 & 58.57 & 53.33 & 97.06 & \multicolumn{1}{>{\columncolor{gdBlueTint}[\tabcolsep][0pt]}r@{}}{43.33} \\
\arrayrulecolor{gdRule}\midrule\arrayrulecolor{black}
\multicolumn{13}{@{}l}{\textcolor{gdGray}{\textit{OOD question types}}} \\
Graph-CoT (3B) & \textcolor{gdGray}{--} & \textcolor{gdGray}{--} & \textcolor{gdGray}{--} & 45.00 & 100.00 & 0.00 & 30.00 & 100.00 & 0.00 & 40.00 & 98.80 & 45.00 \\
Graph Counselor (3B) & \textcolor{gdGray}{--} & \textcolor{gdGray}{--} & \textcolor{gdGray}{--} & \textcolor{gdGray}{--} & \textcolor{gdGray}{--} & 0.00 & \textcolor{gdGray}{--} & \textcolor{gdGray}{--} & 0.00 & \textcolor{gdGray}{--} & \textcolor{gdGray}{--} & 45.00 \\
Vanilla PPO (3B) & \textcolor{gdGray}{--} & \textcolor{gdGray}{--} & \textcolor{gdGray}{--} & 70.00 & 96.48 & 0.00 & 20.00 & 97.01 & 0.00 & 80.00 & 93.51 & 45.00 \\
\rowcolor{gdBlueTint}\multicolumn{1}{@{}>{\columncolor{gdBlueTint}[0pt][\tabcolsep]}l}{\ours{} (3B)} & \textcolor{gdGray}{--} & \textcolor{gdGray}{--} & \textcolor{gdGray}{--} & 95.00 & 98.17 & 0.00 & 90.00 & 100.00 & 0.00 & 90.00 & 100.00 & \multicolumn{1}{>{\columncolor{gdBlueTint}[\tabcolsep][0pt]}r@{}}{45.00} \\
Graph-CoT (14B) & \textcolor{gdGray}{--} & \textcolor{gdGray}{--} & \textcolor{gdGray}{--} & 32.50 & 98.16 & 0.00 & 0.00 & 98.80 & 0.00 & 45.00 & 100.00 & 45.00 \\
\rowcolor{gdBlueTint}\multicolumn{1}{@{}>{\columncolor{gdBlueTint}[0pt][\tabcolsep]}l}{\ours{} (14B)} & \textcolor{gdGray}{--} & \textcolor{gdGray}{--} & \textcolor{gdGray}{--} & 47.50 & 100.00 & 0.00 & 10.00 & 100.00 & 0.00 & 60.00 & 100.00 & \multicolumn{1}{>{\columncolor{gdBlueTint}[\tabcolsep][0pt]}r@{}}{45.00} \\
\bottomrule
\end{tabularx}
\end{table*}

\fi
\begin{table*}[!tp]
\centering
\small
\setlength{\tabcolsep}{4pt}
\caption{Search behavior across all 834 test questions in four mutually
exclusive groups. $n$: questions per group. Calls: mean proposed graph
calls per question. EH: evidence hit (\%). Limit: episodes unfinished at
the interaction limit (\%).}
\label{tab:interaction_complexity}
\begin{tabularx}{\linewidth}{@{}lr*{3}{>{\raggedleft\arraybackslash}X}*{3}{>{\columncolor{gdBlueTint}}>{\raggedleft\arraybackslash}X}*{3}{>{\raggedleft\arraybackslash}X}*{2}{>{\columncolor{gdBlueTint}}>{\raggedleft\arraybackslash}X}>{\columncolor{gdBlueTint}[\tabcolsep][0pt]}>{\raggedleft\arraybackslash}X@{}}
\toprule
 & & \multicolumn{3}{c}{Vanilla PPO (3B)} & \multicolumn{3}{>{\columncolor{gdBlueTint}}c}{\ours{} (3B)} & \multicolumn{3}{c}{Graph-CoT (14B)} & \multicolumn{3}{>{\columncolor{gdBlueTint}[\tabcolsep][0pt]}c@{}}{\ours{} (14B)} \\
\cmidrule(lr){3-5}\cmidrule(lr){6-8}\cmidrule(lr){9-11}\cmidrule(lr){12-14}
Group & $n$ & Calls & EH & Limit & Calls & EH & Limit & Calls & EH & Limit & Calls & EH & Limit \\
\midrule
Easy & 356 & 2.87 & 72.47 & 2.53 & 2.92 & 76.40 & 2.53 & 2.45 & 81.18 & 8.15 & 2.61 & 83.99 & 3.93 \\
Medium & 167 & 5.92 & 28.74 & 20.36 & 6.43 & 37.72 & 25.15 & 5.83 & 49.10 & 38.32 & 5.96 & 57.49 & 28.74 \\
Hard & 241 & 5.10 & 39.00 & 11.62 & 5.49 & 41.91 & 13.28 & 5.89 & 51.04 & 33.61 & 6.44 & 53.53 & 40.66 \\
OOD & 70 & 4.09 & 12.86 & 11.43 & 4.60 & 12.86 & 7.14 & 6.16 & 12.86 & 40.00 & 7.47 & 12.86 & 41.43 \\
\bottomrule
\end{tabularx}
\end{table*}

\begin{table*}[!tp]
\centering
\small
\caption{Question-type breakdown on 92 of the 240 \textsc{Literature}
questions (\%). EM: normalized exact match. R-L: stemmed ROUGE-L.
$n$: questions per type, assigned by ordered matching of question text.}
\label{tab:literature_types}
\begin{tabularx}{\linewidth}{@{}l*{8}{>{\raggedleft\arraybackslash}X}@{}}
\toprule
 & \multicolumn{2}{c}{Simple lookup ($n{=}37$)} & \multicolumn{2}{c}{Aggregation ($n{=}29$)}
 & \multicolumn{2}{c}{Multi-hop ($n{=}21$)} & \multicolumn{2}{c@{}}{Set operations ($n{=}5$)} \\
\cmidrule(lr){2-3}\cmidrule(lr){4-5}\cmidrule(lr){6-7}\cmidrule(l){8-9}
Method & EM & R-L & EM & R-L & EM & R-L & EM & R-L \\
\midrule
Graph-CoT (3B) & 45.95 & 55.62 & 55.17 & 55.17 & 33.33 & 34.38 & 0.00 & 27.43 \\
Graph Counselor (3B) & 48.65 & 55.87 & 48.28 & 48.28 & 28.57 & 29.32 & 0.00 & 16.26 \\
Vanilla PPO (3B) & 51.35 & 56.80 & 58.62 & 58.62 & 33.33 & 39.11 & 20.00 & 58.14 \\
\rowcolor{gdBlueTint}\multicolumn{1}{@{}>{\columncolor{gdBlueTint}[0pt][\tabcolsep]}l}{\ours{} (3B)} & 51.35 & 56.35 & 65.52 & 65.52 & 38.10 & 43.24 & 20.00 & \multicolumn{1}{>{\columncolor{gdBlueTint}[\tabcolsep][0pt]}r@{}}{70.22} \\
\arrayrulecolor{gdRule}\midrule\arrayrulecolor{black}
Graph-CoT (14B) & 51.35 & 58.76 & 72.41 & 72.41 & 42.86 & 42.86 & 60.00 & 61.50 \\
\rowcolor{gdBlueTint}\multicolumn{1}{@{}>{\columncolor{gdBlueTint}[0pt][\tabcolsep]}l}{\ours{} (14B)} & 54.05 & 59.09 & 75.86 & 75.86 & 52.38 & 52.38 & 60.00 & \multicolumn{1}{>{\columncolor{gdBlueTint}[\tabcolsep][0pt]}r@{}}{76.41} \\
\bottomrule
\end{tabularx}
\end{table*}

\ifsubmission\else
\subsection{Behavioral Analysis of Graph Interaction}
\label{sec:appendix_behavior}

To check whether a model follows the multi-round
reasoning--action--observation procedure, we complement the outcome metrics
of Tables~\ref{tab:main} and~\ref{tab:curriculum} with diagnostics computed from execution
traces. Table~\ref{tab:domain_behavior} reports valid-format protocol conformity (VF),
valid-call syntax (VC), and lexical evidence hit (EH) by domain, and
Table~\ref{tab:difficulty_behavior} reports mutually exclusive analysis groups.
VF measures the fraction of episodes that pass the full tag-sequence checker
after common action extraction, which ends each completion at its first
closing graph tag, falling back to its first closing answer tag.
Malformed prefixes are retained, and tags are scored as generated.
VC measures the fraction of proposed calls with an allowed function and the
required basic argument syntax, regardless of whether the node or
relation exists. EH checks executor-returned observations individually
and includes a final tool return at the interaction limit.
For Graph Counselor's multi-agent protocol, only EH applies, computed from
directly recorded tool observations.

Protocol conformity and evidence acquisition can diverge. On
\textsc{Literature} Medium, \ours (3B) mostly follows the protocol, while
the 3B methods rarely retrieve the gold answer in their observations. On
\textsc{Literature} overall, Graph Counselor (3B) has slightly higher EH
than \ours (3B). On \textsc{Literature} Hard, Vanilla PPO (3B) has
slightly higher EH than \ours (3B) at a lower VF.
On OOD question types (Table~\ref{tab:difficulty_behavior}), \ours (3B) has
higher VF than Vanilla PPO (3B) in all three domains, while its OOD R-L in
Table~\ref{tab:difficulty_results} stays low. Within each domain, every
method reaches the same EH on these questions.
\fi

\subsection{Interaction Complexity across Difficulty Levels}

Table~\ref{tab:interaction_complexity} connects the four analysis groups to
the mean number of proposed calls and the fraction of episodes unfinished
at the interaction limit. At 14B, \ours interacts longer on
Hard questions than Graph-CoT, with more proposed calls and
more episodes unfinished at the limit.

\subsection{Question-Type Breakdown on \textsc{Literature}}
\label{sec:lit_qtype}

To localize performance on \textsc{Literature}, Table~\ref{tab:literature_types}
uses a fixed subset of 92 questions, assigned by ordered matching of question
text to simple lookup, aggregation and counting, multi-hop and multiple
constraints, and set operations. Each question enters its first matching
category. All policies use the same subsets. Here normalized EM lowercases answers, removes the
tokens \texttt{a/an/the/usd}, replaces punctuation and underscores with spaces,
and collapses whitespace before exact matching. R-L uses stemming on the
original answers.

Among the 3B rows, \ours has the highest R-L in three of the four
categories, and Vanilla PPO is slightly higher on simple lookup. On
aggregation and multi-hop questions, \ours reaches 65.52 and 43.24 R-L,
compared with 58.62 and 39.11 for Vanilla PPO. At 14B, \ours reaches a
multi-hop R-L of 52.38, compared with 42.86 for Graph-CoT, and the two
methods tie in normalized EM on set operations.

\ifsubmission\else
\subsection{Complete Case-Study Trajectories}
\label{app:full_case_study}

Figure~\ref{fig:case_study_logs} gives both complete trajectories.

\begin{figure*}[tp]
\centering
\begin{tcolorbox}[
    colback=gdGrayTint,
    colframe=gdGray,
    colbacktitle=gdGray,
    coltitle=white,
    boxrule=0.4pt,
    arc=1mm,
    title=Vanilla PPO (3B),
    fonttitle=\bfseries\small,
    boxsep=1mm,
    left=1mm, right=1mm, top=1mm, bottom=1mm
]
\small\ttfamily\raggedright
\textbf{Question:}
What is the usual price of the items often purchased together with Mugen Power 2750mAh Extended Battery for Samsung Galaxy S4 I9500 / I9505?\par
\textcolor{gdGray}{\textbf{[Round 0]}}
\textcolor{gdLavenderDark}{<think>}The question is asking for the usual price of items often purchased together with a specific item, which is Mugen Power 2750mAh Extended Battery for Samsung Galaxy S4 I9500 / I9505. However, the graph structure provided does not contain information about items often purchased together. Therefore, it's not possible to determine the usual price of such items directly from the given graph.\textcolor{gdLavenderDark}{</think>}\par
\textcolor{gdTealDark}{<graph>}NeighbourCheck[Mugen Power 2750mAh Extended Battery for Samsung Galaxy S4 I9500 / I9505, bought\_together\_item]\textcolor{gdTealDark}{</graph>}\par
\textcolor{gdCyanDark}{<information>}The node or neighbor type does not exist in the graph. This might be because your given neighbor type is not correct. Please modify it.\textcolor{gdCyanDark}{</information>}\par
\textcolor{gdGray}{\textbf{[Round 1]}}
\textcolor{gdLavenderDark}{<think>}Since the neighbor type 'bought\_together\_item' does not exist in the graph, it's not possible to determine the usual price of items often purchased together with the given item.\textcolor{gdLavenderDark}{</think>}\par
\textcolor{gdBlue}{<answer>}Not possible to determine\textcolor{gdBlue}{</answer>}
\end{tcolorbox}

\begin{tcolorbox}[
    colback=gdBlueTint,
    colframe=gdBlue,
    colbacktitle=gdBlue,
    coltitle=white,
    boxrule=0.4pt,
    arc=1mm,
    title=\ours{} (3B),
    fonttitle=\bfseries\small,
    boxsep=1mm,
    left=1mm, right=1mm, top=1mm, bottom=1mm
]
\small\ttfamily\raggedright
\textbf{Question:}
What is the usual price of the items often purchased together with Mugen Power 2750mAh Extended Battery for Samsung Galaxy S4 I9500 / I9505?\par
\textcolor{gdGray}{\textbf{[Round 0]}}
\textcolor{gdLavenderDark}{<think>}The question is asking the usual price of items that are often purchased together with a specific item, which is Mugen Power 2750mAh Extended Battery for Samsung Galaxy S4 I9500 / I9505. To find this, I need to first identify the item node for Mugen Power 2750mAh Extended Battery for Samsung Galaxy S4 I9500 / I9505, then find its bought\_together\_item neighbors, and finally check the price feature of these neighbors.\textcolor{gdLavenderDark}{</think>}\par
\textcolor{gdTealDark}{<graph>}RetrieveNode[Mugen Power 2750mAh Extended Battery for Samsung Galaxy S4 I9500 / I9505]\textcolor{gdTealDark}{</graph>}\par
\textcolor{gdCyanDark}{<information>}The ID of this retrieval target node is B00CXOD6BQ.\textcolor{gdCyanDark}{</information>}\par
\textcolor{gdGray}{\textbf{[Round 1]}}
\textcolor{gdLavenderDark}{<think>}The question is asking the usual price of items that are often purchased together with the item with ID B00CXOD6BQ. Next, I need to find the items that are bought together with this item, and then find the price feature of these items.\textcolor{gdLavenderDark}{</think>}\par
\textcolor{gdTealDark}{<graph>}NeighbourCheck[B00CXOD6BQ, bought\_together\_item]\textcolor{gdTealDark}{</graph>}\par
\textcolor{gdCyanDark}{<information>}The bought\_together\_item neighbors of B00CXOD6BQ are: ['B00BW0XZO0'].\textcolor{gdCyanDark}{</information>}\par
\textcolor{gdGray}{\textbf{[Round 2]}}
\textcolor{gdLavenderDark}{<think>}The item B00BW0XZO0 is one of the items often purchased together with the item B00CXOD6BQ. Now I need to find the price feature of this item.\textcolor{gdLavenderDark}{</think>}\par
\textcolor{gdTealDark}{<graph>}NodeFeature[B00BW0XZO0, price]\textcolor{gdTealDark}{</graph>}\par
\textcolor{gdCyanDark}{<information>}The price feature of B00BW0XZO0 are: 44.99.\textcolor{gdCyanDark}{</information>}\par
\textcolor{gdGray}{\textbf{[Round 3]}}
\textcolor{gdLavenderDark}{<think>}The price of the item often purchased together with Mugen Power 2750mAh Extended Battery for Samsung Galaxy S4 I9500 / I9505 is 44.99.\textcolor{gdLavenderDark}{</think>}\par
\textcolor{gdBlue}{<answer>}44.99\textcolor{gdBlue}{</answer>}
\end{tcolorbox}
\caption{Complete trajectories of Vanilla PPO (upper) and \ours (lower)
for Figure~\ref{fig:case_study}. Vanilla PPO makes one graph call and
\ours makes three, answering ``Not possible to determine'' and 44.99
(gold), respectively. Tag colors follow Figure~\ref{fig:framework}.}
\label{fig:case_study_logs}
\Description{Two stacked boxes contain all recorded rounds for Vanilla PPO
(two) and GraphDancer (four), including reasoning, graph calls, observations,
and the final answers ``Not possible to determine'' and 44.99, respectively.}
\end{figure*}

\fi

\section{Ethical Considerations}
\label{app:ethics}

Letting an LLM explore graph-structured knowledge through executable
function calls introduces risks when it connects to real-world graphs.
Graph tool APIs may expose sensitive information, such as proprietary
relations or user data, so deployments should enforce authentication and
authorization, least-privilege function permissions, auditing, and
redaction of sensitive fields. Untrusted text in node attributes and
observations can open systems to prompt injection that
manipulates subsequent actions. Our experiments leave these
safeguards untested. We use an existing public benchmark and collect no new
participant data. An LLM assisted the difficulty labeling of 550
unseen-domain questions (Appendix~\ref{app:labels}), and GPTScore
uses an LLM judge (Section~\ref{sec:exp_setup}). Our biomedical and legal results cover benchmark tasks only, and clinical
or legal decision support lies outside their scope.
Wrong answers and partial evidence remain
possible after training.

\clearpage
\end{document}